\documentclass[lettersize,journal]{IEEEtran}
\usepackage{amsmath,amsfonts}
\usepackage{algorithmic}
\usepackage{algorithm}
\usepackage{array}
\usepackage[caption=false,font=normalsize,labelfont=sf,textfont=sf]{subfig}
\usepackage{textcomp}
\usepackage{stfloats}
\usepackage{url}
\usepackage{verbatim}
\usepackage{graphicx}
\usepackage{cite}
\usepackage{xcolor}
\usepackage{booktabs}
\usepackage[numbers]{natbib}
\usepackage{multirow}
\newcounter{tablerow}
\newcolumntype{N}{>{%
  \ifnum\value{tablerow}=0%
    \setcounter{tablerow}{1}%
    \hspace{4pt}%
  \else
    \thetablerow\hspace{4pt}%
    \stepcounter{tablerow}%
  \fi
}l}

\begin{document}

\title{DyG$^2$T: Modeling Object Dynamics with 3D Gaussian Temporal-Spatial Particle Graph Transformer}

\author{Yansong Wang, Zhaobo Qi, Xinyan Liu, Beichen Zhang, \\ Shuhui Wang, Weigang Zhang, and Qingming Huang,~\IEEEmembership{Fellow, IEEE}


\thanks{Received 1 January 2025; revised 1 January 2025; accepted 1 January 2025. Date of publication 1 January 2025; date of current version 1 January 2025. 
This work was supported in part by the National Natural Science Foundation of China under Grant 62441232, 62306092, and 62476068, and in part by the Natural Science Foundation of Shandong Province, China: ZR2025ZD01, ZR2024QF066, ZR2025MS995. 
This article was recommended by Associate Editor.(\textit{Corresponding author: Weigang Zhang, Zhaobo Qi.})}
\thanks{Yansong Wang, Zhaobo Qi, Xinyan Liu, Beichen Zhang, and Weigang Zhang are with the School of Computer Science and Technology, Harbin Institute of Technology at Weihai, Weihai 264209, China. (e-mail: yansongwang\_hit@stu.hit.edu.cn, \{qizb,xinyliu,beiczhang,wgzhang\}@hit.edu.cn)
}
\thanks{Shuhui Wang is with the Key Laboratory of Intelligent Information Processing, Institute of Computing Technology, Chinese Academy of Sciences, Beijing 100190, China. (e-mail: wangshuhui@ict.ac.cn)}
\thanks{Qingming Huang is with the School of Computer Science and Technology, University of Chinese Academy of Sciences, Beijing 101408, China, and with the Institute of Computing Technology, Chinese Academy of Sciences, Beijing 100190, China. (e-mail: qmhuang@ucas.ac.cn)}
}

\markboth{IEEE Transactions on Circuits and Systems for Video Technology,~Vol.~xx, No.~x, xxx~2025}%
{Wang \MakeLowercase{\textit{et al.}}: DyG$^2$T: 3D Gaussian Temporal-Spatial Particle Graph Transformer}


\maketitle

\begin{abstract}
Modeling object dynamics from limited visual observations is a fundamental problem for enabling accurate motion trajectory prediction in embodied interaction scenarios. Existing dynamics modeling methods first compress reconstructed particle representations into sparse Key Points and model their evolution using locally constrained interactions, thereby discarding fine-grained local details and obscuring discriminative interaction modeling across spatial and temporal scales, leading to drifting trajectories and inaccurate appearance prediction. To tackle these issues, we propose DyG$^2$T, a dynamics modeling framework that infers object motion trajectories by spatially completing and temporally discriminating Key Point representations and modeling multi-scale interaction over particle graphs. Spatially, DyG$^2$T enriches each Key Point by aggregating neighboring raw particle positions to recover fine-grained local details, while explicitly encoding relative offsets among Key Points to enhance geometric structure perception. Temporally, we introduce a Temporal Disentangling Network (TDN) to identify dominant cross-frame variations in latent space and amplify inter-frame differences, yielding temporally discriminative representations that are subsequently aggregated via Temporal Attention to capture frame-wise temporal evolution cues. For comprehensive interaction modeling, a Particle Graph Transformer leverages global attention to preserve discriminative long-range dependencies among Key Points, mitigating representation homogenization induced by locality-constrained modeling and providing a robust basis for accurate trajectory prediction. Experiments on both synthetic and real-world datasets demonstrate that DyG$^2$T achieves accurate dynamics modeling and reasoning, and exhibits strong cross-object and real-world generalization. 
\end{abstract}


\section{Introduction}
\IEEEPARstart{R}{ecently}, artificial intelligence has evolved from disembodied intelligence operating independently of the physical world toward embodied systems that require interaction with more realistic, complex real-world environments~\cite{zhang2025catch}. Under this paradigm shift, dynamics modeling and reasoning for dynamic objects have become increasingly important, as applications such as embodied intelligence, digital twins, and interactive simulation require not only understanding the current state of objects but also inferring their future motion from limited visual observations. Interaction with elastic objects represents one particularly challenging scenario, since such objects exhibit complex material-dependent dynamic behaviors, while their internal motion patterns are difficult to fully observe from sparse viewpoints. Consequently, object dynamics modeling has emerged as a fundamental technical demand for capturing motion patterns in continuous dynamic spaces by modeling complex interactions among internal components, such as cross-particle influence propagation~\cite{zhong2024reconstruction}, ultimately enabling precise future trajectory prediction to support seamless agent–environment interaction.

To acquire object dynamics, most existing approaches~\cite{cai2024gic,zhang2025dynamic} first reconstruct 3D particle representations of motion objects from 2D visual observations~\cite{kerbl20233d,li2023hybrid}, then downsample them via Farthest Point Sampling (FPS)~\cite{qi2017pointnet++} to obtain sparse Key Points for dynamic modeling. Current methods largely fall into two categories. Strong physics-prior approaches~\cite{li2023pac,liu2025unleashing} assume a priori of material properties~\cite{cai2024gic} and select corresponding differential equations to evolve Key Points trajectories~\cite{xie2024physgaussian}. However, such reliance on constitutive models restricts their applicability to objects with unknown or heterogeneous materials~\cite{mittal2025uniphy}. In contrast, differentiable network-based methods~\cite{zhao2024gaussianprediction,furecon} learn dynamics in a data-driven manner, avoiding material priors and handcrafted equations. They typically use multilayer perceptrons (MLPs) to infer cross-frame deformations from Key Point spatial attributes, enabling evolution from a canonical representation to arbitrary time steps~\cite{sun20243dgstream,yang2024deformable}, or employ graph neural networks (GNNs)~\cite{zhong2024reconstruction, zhang2025dynamic} to learn Key Point motion patterns within local neighborhoods for future trajectory inference. Yet, relying solely on Key Point-level attributes discards the rich local details of the raw particles, while locality-constrained message passing often leads to homogenized features. These limitations ultimately result in inaccurate appearance predictions and drifting trajectories.

\begin{figure*}[t]
\centering
\includegraphics[width=0.9\textwidth, trim=0 0 0 20]{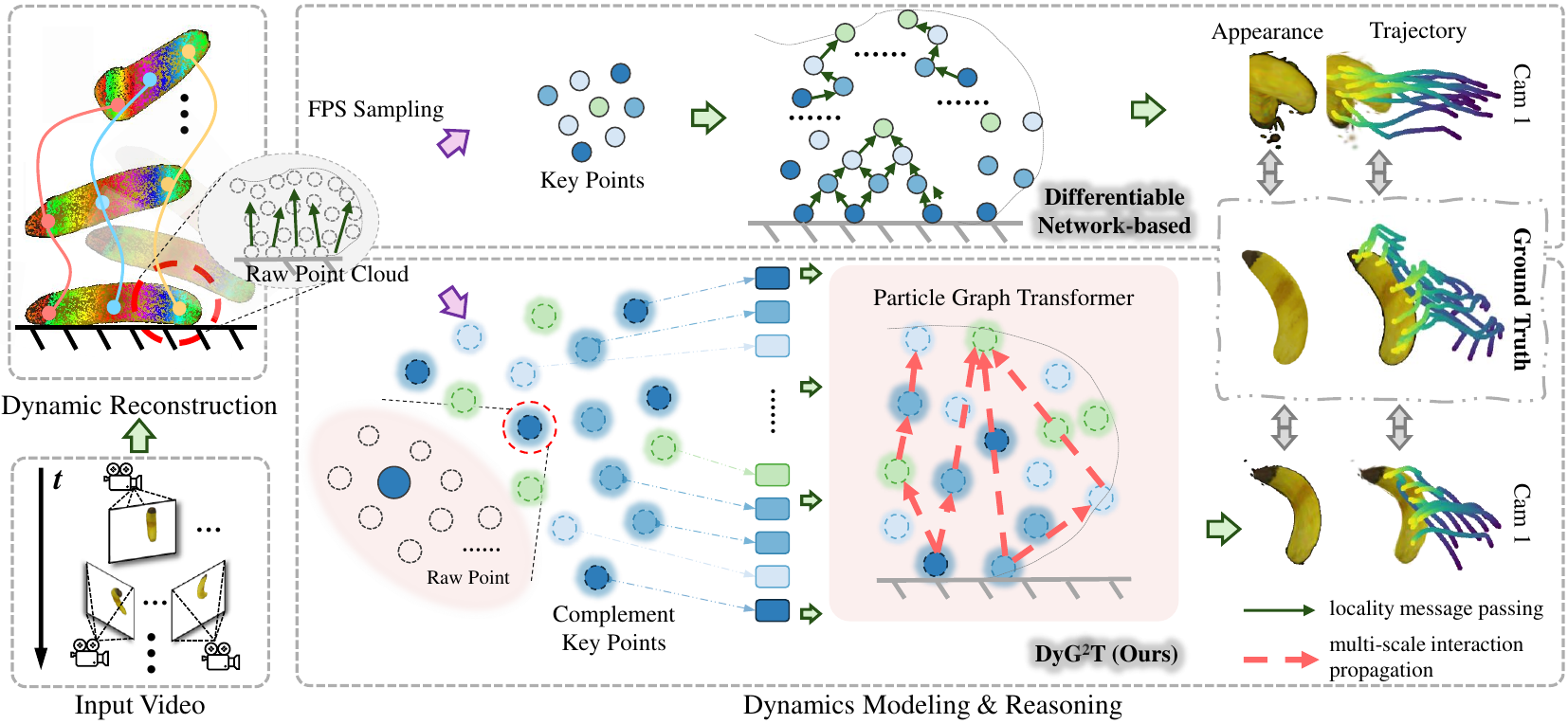}
\caption{DyG$^2$T vs. differentiable network-based methods for dynamics modeling of a falling Banana. Colored curves denote motion trajectories. Differentiable network-based methods rely on Key Point-level spatial attributes and propagate features node-wise, leading to appearance artifacts and trajectory drift. In contrast, DyG$^2$T constructs discriminative embeddings by modeling multi-scale interaction based on completed particle features, enabling accurate trajectory and appearance decoding.}
\label{fig1}
\vspace{-1em}
\end{figure*}

Specifically, in one respect, researchers adopt the coordinates of downsampled Key Points as the initial particle representation for dynamics modeling. As shown in Figure~\ref{fig1}, FPS drops a significant amount of raw particles ({\it i.e.}, Raw Point Cloud), consequently losing fine-grained local details carried by those discarded points; moreover, relying on coordinates as initial features further weakens the model’s perception of object geometry and structure. This dual loss of local details and geometric cues compromises the integrity of the feature basis. Such incomplete particle representations from different frames become temporally entangled, causing semantic collapse that severely reduces temporal discriminability and hinders downstream dynamics modeling, leading to errors in appearance and trajectory decoding. Therefore, retaining the particle representations that are both complete and temporally discriminative is crucial to robust dynamics modeling.

In another respect, locality-constrained message passing limits particle representation learning by inducing feature homogenization. As shown in Figure~\ref{fig1} top, non-local information is obtained only through iterative neighbor aggregation, leading to diffusive updates that progressively blur representational distinctions~\cite{sun2022mhnf}. This homogenization obscures interaction patterns across different spatial extents, producing homogenization particle embeddings and resulting in positional decoding errors and trajectory deviations. Recent studies~\cite{sun2025gtc} suggest that capturing dependencies across multiple spatial ranges explicitly provides an effective means to preserve discriminative particle representations. By aggregating interaction cues spanning different spatial ranges, multi-scale modeling preserves informative variations in particle embeddings. This, in turn, provides a more reliable representational basis for accurate dynamics modeling and trajectory prediction.

To address these issues, we propose DyG$^2$T, a dynamics modeling framework that enables accurate motion inference by spatially completing and temporally disentangling enriched Key Point representations, while explicitly preserving particle feature diversity through the capture of discriminative cross-scale dependencies. As shown in Figure~\ref{fig1}, DyG$^2$T first leverages dynamic reconstruction to extract trackable particle representations from sparse-view videos. To ensure complete spatiotemporal representations, we enhance particle features from two complementary perspectives. Spatially, each Key Point aggregates neighboring raw particle positions that encode local appearance, enriching fine-grained local details. Relative offsets capturing directed layout relations between Key Points are further used to enhance geometric structure perception explicitly. Temporally, we feed adjacent-frame Key Point representations into a Temporal Disentangling Network (TDN) to estimate offsets along dominant disentangling directions, amplifying inter-frame differences and yielding temporally discriminative features. These disentangled embeddings are then aggregated via Temporal Attention to enrich Key Points' frame-wise temporal evolution cues. To further model multi-scale interaction from local contacts to global structure, a Particle Graph Transformer leverages global attention to capture multi-scale dependencies among interaction-relevant Key Points directly, preserving discriminative features essential for accurate dynamics modeling.

We evaluate DyG$^2$T on the Spring-Gaus synthetic \& real-world~\cite{zhong2024reconstruction} and our Unity3D-Heterogeneous datasets, demonstrating its strong cross-object generalization and real-world scalability in dynamics modeling and reasoning. The contributions of this work are as follows:

\begin{itemize}
\item We propose DyG$^2$T, a dynamics modeling framework that mitigates trajectory prediction bias by capturing differentiated multi-scale dependencies within particle representations. 

\item DyG$^2$T enriches particle representations with fine-grained local details and geometry construction through spatial completion, and captures discriminative temporal evolution cues via TDN and temporal aggregation.

\item Extensive experiments on both synthetic and real-world datasets demonstrate that DyG$^2$T accurately predicts object motion trajectories across diverse geometries and scales effectively in real-world scenarios.
\end{itemize}

\section{Related Work}
\subsection{Dynamic 3D Scene Reconstruction}
Reconstructing 3D dynamic scenes from multi-view images has become a central approach for acquiring 3D representations of dynamic scenes~\cite{chen2025g3flow}. Existing methods can be broadly grouped into dynamic extensions of neural radiance fields (NeRFs)~\cite{pumarola2021d,li2023representing} or 3D Gaussians~\cite{furecon,liang2025himor}. Early efforts extend NeRF representations~\cite{mildenhall2021nerf,wang2021neus}, originally designed for static scenes, to dynamic settings by either fitting frame-wise residual motion~\cite{li2022neural,wang2023neus2} or learning deformation fields that map a canonical template to each observed frame~\cite{park2021nerfies,gao2022monocular}. While these approaches can reconstruct 3D geometry, their ray-sampling nature leads to slow training~\cite{ding2024ray}, and the purely data-driven implicit representations exhibit limited generalization and poor robustness to noise, ultimately constraining downstream dynamics modeling and prediction.

In contrast, 3D Gaussian-based methods build upon 3D Gaussian Splatting (3DGS)~\cite{kerbl20233d}, representing dynamic objects as explicit Gaussian spheres or ellipsoids. This explicit parameterization enhances controllability and provides a solid foundation for dynamic extensions. Dyn3DGS~\cite{luiten2024dynamic}, the first dynamic extension of 3DGS, optimizes non-visual Gaussian parameters per frame to enable trackable dynamic reconstruction. Similar to NeRF-based deformation fields, subsequent work~\cite{li2025frpgs,guo2024motion,yang2024deformable} introduces MLP-based deformation networks to query temporally varying Gaussian attributes, enabling continuous warping from a canonical representation to arbitrary observations. Follow-up methods expand this direction: Scale-GS~\cite{yang2025scale} and SCas4D~\cite{lyu2025scas4d} integrate hierarchical clustering into the deformation network to achieve layered fitting of dynamic Gaussians, while 3DGStream~\cite{sun20243dgstream} decodes Gaussian rotations and translations from hashed voxel features. Distinct from these reconstruction-focused approaches that do not account for physical behavior, our work incorporates dynamics modeling to support future motion inference.

\subsection{Physics-based Dynamics Modeling and Reasoning}

Inferring object dynamics from only sparse initial observations is highly challenging due to the complexity of continuous state spaces~\cite{cai2024gic}. Existing dynamics modeling and reasoning methods can be broadly divided into 2 groups, including strong physics-prior approaches, such as physics-engine-based and PDE-regularized methods, and differentiable network-based methods. Physics-engine-based approaches~\cite{liu2025unleashing} treat material properties as priors and evolve trajectories using constitutive equations corresponding to each material. Recent studies~\cite{cai2024gic,mittal2025uniphy} estimate physical parameters in a reverse manner to reduce reliance on predefined material settings, yet they remain constrained by the limited generalizability of specific constitutive formulations. Inspired by Physics-Informed Neural Networks (PINNs)~\cite{raissi2019physics}, PDE-regularized methods~\cite{im2024regularizing,raissi2020hidden,wang2024physics} incorporate various PDEs as soft constraints to regularize the learning of physical properties of 3D objects, but they often suffer from low training efficiency and difficulty in modeling complex boundary conditions~\cite{li2025freegave}.

\begin{figure*}[t]
\centering
\includegraphics[width=\textwidth, trim=0 0 0 0]{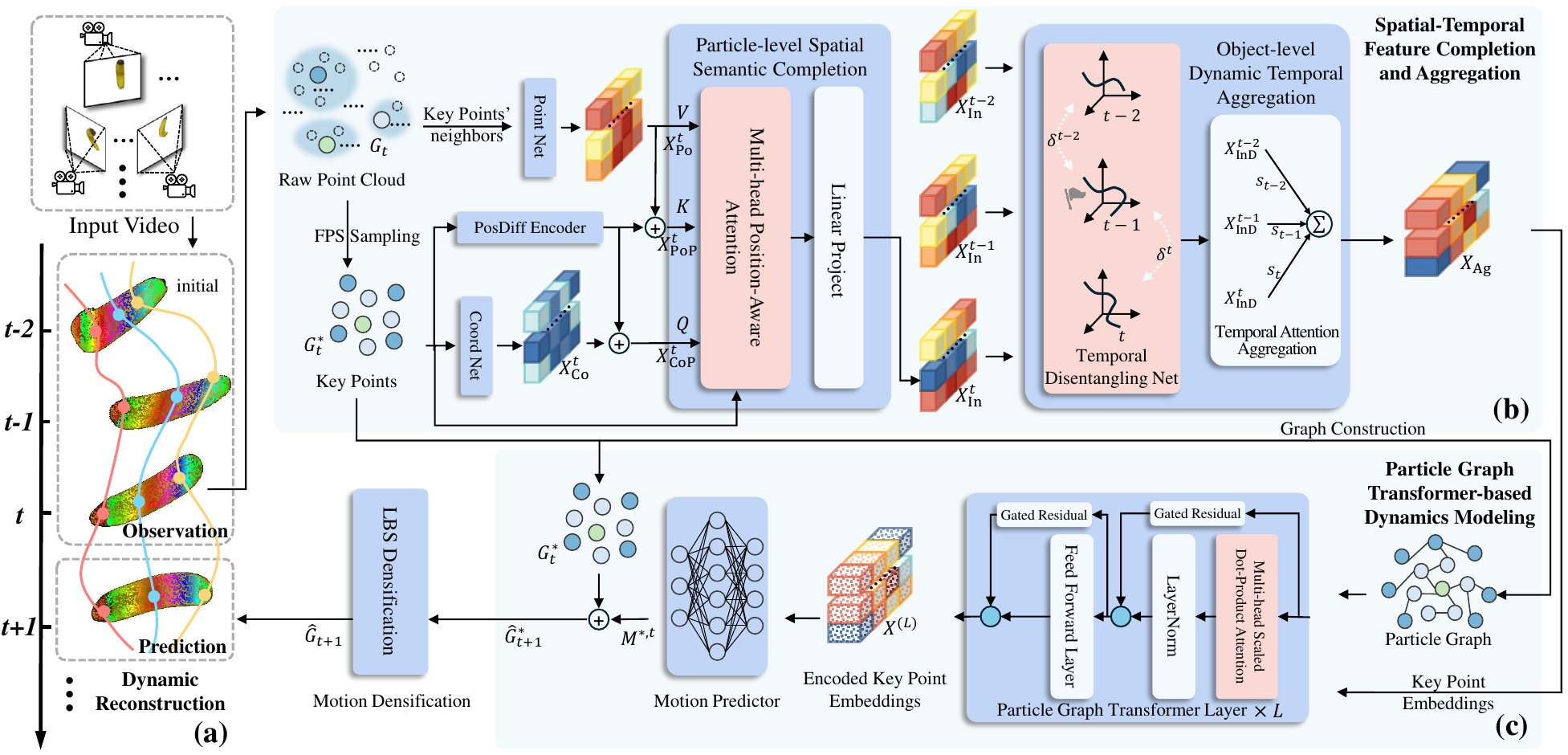}
\caption{
Overview of the DyG$^2$T. (a) DyG$^2$T utilizes dynamic reconstruction to extract trackable particle representations. (b) The Spatial-Temporal Feature Completion and Aggregation performs semantic completion and temporal aggregation at both particle and object levels, which spatiotemporally enhance particle representations. (c) With the global attention of the Particle Graph Transformer, DyG$^2$T enables discriminative multi-scale dependencies modeling.}
\label{fig2}
\vspace{-1em}
\end{figure*}

In contrast, differentiable-network-based approaches shift the burden of modeling complex interactions to nonlinear architectures such as MLPs~\cite{li2025freegave,hu2025learnable}, GNNs~\cite{li2019learning,zhang2025dynamic}, or spring-mass~\cite{zhong2024reconstruction}, achieving strong results on simulated or real entities, including animals~\cite{li2025freegave}, cloth~\cite{lin2022learning}, fluids~\cite{li2019learning}, and elastic materials~\cite{zhong2024reconstruction}. FreeGave~\cite{li2025freegave} focuses on joint novel view synthesis and future prediction from free-view video, but its divergence-free deformation field, based on MLPs, neglects discontinuous state changes induced by collisions and other dynamic interactions. Methods such as GaussianPrediction~\cite{zhao2024gaussianprediction} instead treat downsampled Key Points as nodes and model dynamics via local message passing, which inevitably discards fine-grained geometric details and overlooks interaction modeling across different spatial scales.

In comparison, our method lies between these directions: it combines physics-regularized dynamic reconstruction with differentiable dynamics prediction, and constructs a Spatial-Temporal Graph Transformer that directly captures object-internal multi-scale interaction patterns induced by object motion, striking a balance between physical plausibility and generalization to unknown or heterogeneous materials.

\section{Method}

\subsection{Preliminary: 3D Gaussian Spaltting}\label{sec_preliminary}

3D Gaussian Splatting (3DGS)~\cite{kerbl20233d} optimizes a set of learnable Gaussian kernels as an explicit dynamic object representation through a differentiable rasterizer. Each kernel is parameterized by spatial and appearance descriptors. The spatial descriptors include the 3D position $\mu=(x, y, z)$ and the 3D rotation $r=\left(qw, qx, qy, qz\right)$ represented as a quaternion, where $qw$ is the quaternion real part, and $(qx, qy, qz)$ is the quaternion imaginary components. The appearance descriptor includes the 3D scaling factor $s\in \mathbb{R}^3$, m-DoF spherical harmonics coefficients $\mathbf{h}\in \mathbb{R}^{3(m+1)^2}$, and opacity $\sigma\in [0,1]$. The color $\mathbf{C}$ of each 2D pixel is computed using a depth-sorted Max Volume Rendering~\cite{luiten2024dynamic}:
\begin{equation}\small
    \mathbf{C}=\sum_{i\in N}c_i\phi_i^{2D}\prod_{j=1}^{i-1}(1-\phi_j^{2D}),
\end{equation}
where $N$ denotes the set of the Gaussian kernels, and $c_i$ is the RGB color of kernel $i$ obtained from spherical harmonics based on the view direction and coefficients $\mathbf{h}$. $\phi_i$ is the weighted opacity of kernel $i$,
\begin{equation}\label{eq2}\small
    \phi_i=\sigma_i\exp\left(-\frac{1}{2}(\mathbf{x}-\mu_i)^{\mathsf{T}}\Sigma_i^{-1}(\mathbf{x}-\mu_i)\right),
\end{equation}
where $\Sigma_i=R_is_is_i^{\mathsf{T}}R_i^{\mathsf{T}}$ denotes the covariance matrix of Gaussian $i$ composed of a scaling matrix $s_i=\mathrm{diag}(sx_{i}, sy_{i}, sz_{i})$ and a rotation matrix $R_{i}= \mathrm{q2R}(qw_i, qx_i, qy_i, qz_i)$. $\mathrm{q2R}(\cdot)$ converts a quaternion into a rotation matrix. $\phi^{2D}_i$ is the 2D version of Equation~\ref{eq2}. The pixel position is obtained by approximating a perspective projection of the 3D Gaussian's position $\mu_{i}^{2D}$ and covariance matrix $\Sigma_{i}^{2D}$:
\begin{equation}\small
\begin{split}
    \mu_{i}^{2D}&=(P((E\mu_{i})/(E\mu_{i})_{z}))_{1:2}\\
    \Sigma_{i}^{2D}&=(JE\Sigma_{i}E^{\mathsf{T}}J^{\mathsf{T}})_{1:2,1:2}
\end{split},
\end{equation}
where $E$ and $P$ represent the extrinsic and intrinsic parameters of the view camera, respectively, and $J$ is the Jacobian matrix of the projection transformation.

\subsection{Overivew}
For each dynamic object, given a set of 2D observations $\left\{I_{o,1},\ldots,I_{o,t} \right\}_{o=1}^{O}$ captured from $O$ views over $t$ timesteps, our objective is to predict the object’s motion over the subsequent $\epsilon$ timesteps via dynamics modeling. For clarity, we capture one frame per timestep and refer to the $O \times t$ frames from $1$ to $t$ as observation frames, where $t=1$ denotes the initial frame. The subsequent $\epsilon$ frames are referred to as prediction frames.

As the data foundation for dynamics modeling, we reconstruct the trackable raw particle sequences from 2D visual observations for the current object (as shown in Figure~\ref{fig2}(a)). Specifically, we first apply vanilla 3DGS~\cite{kerbl20233d} on the initial frame $\left\{I_{o,1}\right\}_{o=1}^O$ to obtain the initial 3D Gaussian parameters. For subsequent observation frames, we follow Dyn3DGS~\citep{luiten2024dynamic} to perform frame-wise reconstruction with trackable Gaussians. Appearance descriptors are frozen and reused across frames, as they encode view-dependent object appearance that remains approximately consistent over short temporal intervals. Spatial descriptors are iteratively optimized under multi-view constraints $\left\{I_{o,2},\ldots, I_{o,t}\right\}_{o=1}^O$. Since Dyn3DGS initializes each frame from the previous one, this process naturally yields temporally trackable spatial descriptors, forming raw particle 3D position sequences $\left\{G_1, \dots, G_t\right\}$, where $G_t=\left\{\mu_i^t\middle|i \in [1, N]\right\}$, along with corresponding 3D rotation sequences. For notational convenience, we also refer to parameterized 3D Gaussians as particles.

To predict the future motion of objects parameterized by 3D Gaussians, we design a Spatial-Temporal Feature Completion and Aggregation module and a Particle Graph Transformer-based Dynamics Modeling module. As shown in Figure~\ref{fig2}(b), we first apply FPS to sample Key Points from the raw particle sequence ({\it i.e.}, the Raw Point Cloud). Since subsequent dynamics modeling mainly operates on 3D positions, we represent the downsampled Key Points simply by their 3D position sequences $\{G_1^\ast, \ldots, G_t^\ast\}$, where $G_t^\ast=\{\mu_i^{\ast,t}\mid i \in [1, N^\ast]\}$. The Spatial-Temporal Feature Completion and Aggregation mechanism subsequently operates on Raw \& Key Points to perform particle-level spatial semantic completion and object-level dynamic temporal aggregation, resulting in enriched Key Point embeddings $X_{\text{Ag}}$. We then employ the Particle Graph Transformer to construct direct interaction pathways between Key Points, enabling accurate trajectory modeling and precise displacement prediction $M^{\ast,t}$. Afterwards, the positions of the Key Points $\hat{G}^\ast_{t+1}$ at frame $t+1$ is updated by adding predicted displacement $M^{\ast,t}$ to $G^\ast_{t}$ (Figure~\ref{fig2}(c)). Finally, using Linear Blend Skinning (LBS)~\cite{sumner2007embedded, huang2024sc}, we interpolate the predicted Key Point positions $\hat{G}^\ast_{t+1}$ to estimate the 3D positions $\hat{G}_{t+1}$ and 3D rotations of raw particles for the next frame. Appearance descriptors are directly reused from the previous frame. This yields the complete 3D Gaussian parameters for future frames. By iteratively incorporating predicted frames into the observation frames, we enable motion prediction over the subsequent $\epsilon$ frames. In practice, the range of observation frames is set to 3 ({\it i.e.}, from $t-2$ to $t$).

\subsection{Spatial-Temporal Feature Completion and Aggregation}\label{sec32}

In this section, we enhance particle representations from two complementary perspectives: spatial semantic completion and dynamic temporal aggregation.

As a prerequisite, we repeatedly apply Farthest Point Sampling (FPS)~\cite{eldar1997farthest} to extract a sequence of Key Points  $\{G_1^\ast, \ldots, G_t^\ast\}$ from the Raw Point Cloud, where FPS iteratively selects points that are farthest from the already sampled set to ensure uniform spatial coverage. Based on the downsampled Key Point sequence, as illustrated in Figure~\ref{fig2}(b), we first introduce a Particle-level Spatial Semantic Completion module, which enriches each Key Point with fine-grained local details and geometric structure by jointly integrating neighborhood positional information from the Raw Point Cloud and relative offset relationships among Key Points. Subsequently, an Object-level Dynamic Temporal Aggregation module is employed to model temporal dynamics, in which a Temporal Disentangling Net (TDN) estimates dominant cross-frame disentangling offsets in the latent space to amplify inter-frame differences, followed by a Temporal Attention mechanism that selectively aggregates discriminative features to capture frame-wise temporal evolution cues.

\subsubsection{Particle-level Spatial Semantic Completion}

This module leverages a Position-Aware Attention mechanism to enrich the initial coordinate-level representations by aggregating positional information from neighboring raw particles and the relative offsets among Key Points. 

As shown in Figure~\ref{fig2}(b), to obtain the coordinate-level initial representation, we first employ Coord Net, which is the 2-layer MLPs with ReLU, to map the Key Point 3D position $\mu_i^{\ast,t} \in G_t^\ast$ into coordinate features $X_{\text{Co}}^t \in \mathbb{R}^{N^\ast \times H_{\text{Co}}}$. 

Then, we employ PointNet~\cite{qi2017pointnet} to encode the positions of the $k$-nearest Raw Points for each Key Point, map them into a unified feature space using a shared MLP, and apply max pooling over the $k$ neighbors to obtain the neighborhood feature $X_{\text{Po},i}^t \in \mathbb{R}^{H_{\text{Po}}}$. Here, max-pooling serves as a neighborhood aggregation mechanism that preserves the most salient local responses while suppressing potentially noisy information, thereby aggregating permutation-invariant fine-grained local appearance information from the unordered Raw-Point positional neighborhood, ultimately enhancing the spatial representation capability of each Key Point.

Subsequently, we employ a PosDiff Encoder that models the pairwise relative offset between Key Points to capture the directed layout relations of Key Points. In detail, we encode the coordinate differences between Key Points using a 2-layer MLP with ReLU. The pairwise encodings are reduced to nodewise via neighbor-wise mean aggregation and get relative geometric embeddings for Key Points.

Afterwards, we add these relative geometric embeddings to the coordinate features $X_{\text{Co}}^t$ and neighborhood features $X_{\text{Po}}^t$, yielding enhanced coordinate and neighborhood embeddings, $X_{\text{CoP}}^t \in \mathbb{R}^{N^\ast \times H_{\text{Co}}}$ and $X_{\text{PoP}}^t \in \mathbb{R}^{N^\ast \times H_{\text{Po}}}$, respectively. By injecting identical relative offset cues into both feature streams, the PosDiff Encoder establishes a shared relational reference between complementary semantics that enables consistent reasoning under directed Key Point interactions. In addition, this symmetric enhancement is particularly important for subsequent interaction modeling, where feature updates between Key Points are inherently direction-dependent~\cite{zhao2021point}~(e.g., information propagated A$\to$B may differ from B$\to$A). Encoding relative geometry beforehand ensures that such bidirectional interactions are grounded in explicit geometric cues, thereby improving both local detail and structural preservation.

Finally, we design a Multi-head Position-Aware Attention mechanism that enriches the Key Point enhanced coordinate features $X_{\text{CoP}}^t$ by aggregating the neighborhood representations $X_{\text{Po}}^t$ and the PosDiff-enhanced neighborhood features $X_{\text{PoP}}^t$, allowing the coordinate embeddings to incorporate fine-grained local detail and relative layout information. Meanwhile, we incorporate a Gaussian-weighted relation prior $\omega$ to regularize the attention distribution~\cite{yang2025trans}, ensuring that it adheres to plausible geometric correlations among Key Points. For clarity, we illustrate the procedure using the Single-Head formulation,
\begin{equation}
    Q=X_{\text{CoP}}^t\mathbf{W}^Q,K=X_{\text{PoP}}^t\mathbf{W}^K,V=X_{\text{Po}}^t\mathbf{W}^V,
\end{equation}
\begin{equation}
    X_{\text{In}}^{t}=\text{SoftMax}\left(\frac{Q\cdot K^\mathsf{T}}{\sqrt{H_K}}+\log{\left(\omega\right)}\right)V,
\end{equation}
where $\mathbf{W}^Q, \mathbf{W}^K \in \mathbb{R}^{H_{\text{Co}} \times H_K}$ and $\mathbf{W}^V \in \mathbb{R}^{H_{\text{Po}} \times H_V}$ are learnable matrices, $\omega = \exp\left(-\frac{d^2}{2\rho^2}\right)$ is Gaussian-weighted relation, $d$ is the Euclidean distance between Key Points and $\rho$ is a sharpness coefficient. $X_{\text{In}}^{t} \in \mathbb{R}^{N^\ast \times H_{\text{In}}}$ is the spatial semantic features at frame $t$. Applying the same procedure to Key Points at frame $t-2$ and $t-1$, we obtain the spatial semantic feature sequence $\{X_{\text{In}}^{t-2}, X_{\text{In}}^{t-1}, X_{\text{In}}^t\}$.


\subsubsection{Object-level Dynamic Temporal Aggregation}

To enrich the particle embeddings with meaningful dynamic evolutions, we aggregate spatial semantic features across frames. However, directly aggregating multi-frame features without explicit temporal disentanglement often leads to severe semantic collapse: features from different frames become highly overlapped in distribution. This phenomenon is observed in our visualizations (see Figure~\ref{fig_ablationTDN}) and is aligned with theoretical observations that~\cite{zhangdoes,tian2021understanding}, in the absence of explicit diversity or correspondence constraints, the optimal solution of a representation learner tends to degenerate into an almost constant embedding that maps all inputs to the same region of the latent space. For dynamic modeling, such a collapse drastically impairs temporal discriminability and prevents downstream modules from capturing meaningful temporal information.

To address this issue, building on the strong feature foundation provided by spatial completion, we design a Temporal Disentangling Network (TDN) to reconstruct inter-frame differences in cross-frame representations before temporal aggregation, yielding temporally discriminative embeddings. We intentionally adopt a 3-frame observation window to encourage TDN to disentangle short-range inter-frame variations. This local design constrains the disentangling process within a relatively stable temporal neighborhood, thereby avoiding the large inter-frame distribution discrepancies and instability introduced by long-range temporal alignment. To provide a stable canonical anchor for disentangling offset estimation and temporal difference reconstruction, we first designate frame $t-1$ as the reference frame and enforce zero offset, i.e., $X_{\text{InD}}^{t-1} = X_{\text{In}}^{t-1}$. This symmetric center-frame reference design further alleviates the potential bias that may arise when using a one-sided endpoint frame as the temporal reference. We then concatenate the observed frames' spatial semantic features, including the reference frame, to form a temporal joint representation $X_{\text{InC}}$,
\begin{equation}
    X_{\text{InC}} = \text{Concat}\left(X_{\text{In}}^{t-2}, X_{\text{In}}^{t-1}, X_{\text{In}}^t\right) \in \mathbb{R}^{N^\ast \times 3H_{\text{In}}},
\end{equation}
and passed through a linear layer with $\tanh$ activation to estimate the disentangling offsets $\delta^{t-2},~\delta^{t}$ of the non-reference frames $t-2,~t$ relative to the reference frame $t-1$,
\begin{equation}
    \delta^{t-2}, \delta^{t}=\tanh\left({\text{Linear}\left(X_{\text{InC}}\right)}\right)\in \mathbb{R}^{N^\ast \times H_{\text{In}}}.
\end{equation}

These offsets represent the dominant disentangling directions of adjacent-frame spatial semantic representations. By adding the offsets to the spatial features, we pull non-reference representations along these directions, separating them from the entangled spatial features and amplifying inter-frame differences, thereby obtaining temporally discriminative disentangled features $X_{\text{InD}}^{t-2},~X_{\text{InD}}^{t}$,
\begin{equation}
    X_{\text{InD}}^{t-2}=X_{\text{In}}^{t-2}+\delta^{t-2},~X_{\text{InD}}^{t}=X_{\text{In}}^{t}+\delta^{t}.
\end{equation}

Subsequently, we apply a Temporal Attention module to aggregate the disentangled Key Point features across frames adaptively, capturing frame-wise temporal evolution cues. This mechanism assigns frame-wise importance scores $s_i$ based on a nonlinear transformation, enabling the model to emphasize temporally informative cues while suppressing redundant ones,
\begin{equation}
X_{\text{Ag}}=\sum_{i=t-2}^{t}\frac{\exp{\left(s_i\right)}}{\sum_{j=t-2}^{t}\exp{\left(s_j\right)}}X_{\text{InD}}^i,
\end{equation}
\begin{equation}
s_i=\text{Linear}\left(\tanh{\left(\text{Linear}\left(X_{\text{InD}}^i\right)\right)}\right),
\end{equation}
where the scalar scores $s_i$ encode the temporal saliency of each frame, and the resulting aggregated representation $X_{\text{Ag}} \in \mathbb{R}^{N^\ast \times H_{\text{Ag}}}$ captures the dynamically weighted fusion of the temporally disentangled features.

\subsection{Dynamics Modeling Based on Particle Graph Transformer}\label{sec33}

In this section, we describe how the Particle Graph Transformer is utilized to capture multi-scale interaction patterns for preserving discriminative particle-wise information, and to predict the translation vectors $M^{\ast,t} \in \mathbb{R}^{N^\ast \times 3}$ of Key Points at frame $t$, estimating the position $\hat{G}_{t+1}^\ast$ at frame $t+1$. 

As shown in Figure~\ref{fig2}(c), inspired by~\cite{shi2021masked}, we introduce a Particle Graph Transformer. At each iteration step, we construct a Particle Graph from the current Key Point positions by connecting each Key Point to its top-$k_G$ nearest neighbors within a distance threshold $d_e$. The resulting binary adjacency is vectorized and encoded as the learnable edge embedding $e_{ij}$ between nodes $i$ and $j$. After each Key Point position update, the graph is reconstructed using the newly predicted positions. The pseudocode for the Particle Graph construction and evolution process is provided in the Supplementary Material. We then perform global interaction modeling through Graph Attention. Specifically, the aggregated Key Point features $X_{\text{Ag}}$ are projected into the key $k^{(0)}\in\mathbb{R}^{N^\ast\times d_k}$, query $q^{(0)}\in\mathbb{R}^{N^\ast\times d_q}$, and value $v^{(0)}\in\mathbb{R}^{N^\ast\times d_v}$ using separate linear layers. The attention scores $\alpha$ are computed using a scaled dot-product function:
\begin{equation}
    \alpha_{ij}^{(l)}=\frac{\langle q_{i}^{(l)},k_{j}^{(l)}+e_{ij}\rangle}{\sum_{u\in\mathcal{N}(i)}\langle q_{i}^{(l)},k_{u}^{(l)}+e_{iu}\rangle}, 
\end{equation}
where $\langle q,k\rangle=\exp\left(\frac{q^{\mathsf{T}}k}{\sqrt{d_k}}\right)$, $i$ and $j$ are the endpoints of an edge, $i \in N^\ast$, $\mathcal{N}$ represents the set of neighbors of node $i$, and $l = 1, 2, \ldots, L$ denotes the index of the Particle Graph Transformer layer. Finally, we selectively aggregate node features over the entire graph to obtain $X^{(l)} \in \mathbb{R}^{N^\ast \times H_\text{G}}$:
\begin{equation}
    X^{(l)}=\mathop{\sum\sum}_{i\in N^\ast~j\in \mathcal{N}(i)}\alpha_{i,j}^{(l)}(v_{j}^{(l)}+e_{i,j}).
\end{equation}

The Particle Graph Transformer leverages global attention over the particle graph to directly model long-range dependencies that are inaccessible to purely local message passing. By doing so, it mitigates representation homogenization and preserves discriminative particle-wise features, providing a robust representational basis for accurately decoding complex motion trajectories. In addition, to alleviate feature homogenization that may arise during multi-scale message passing, we introduce Gated Residual Connections between successive transformer layers to retain informative signals selectively while preventing over-smoothing. Further architectural details are provided in the Supplementary Material.

Finally, the output representation $X^{(L)}$ from the Particle Graph Transformer is fed into a 3-layer MLP with ReLU to decode the Key Point displacement field $M^{\ast,t} \in \mathbb{R}^{N^\ast \times 3}$. This displacement field provides a direct parametric estimate of the particle-wise motion induced by the underlying dynamics. The predicted Key Point positions at frame $t+1$ are then obtained via simple forward integration:
\begin{equation}
\begin{split}
    \hat{G}_{t+1}^\ast &= \left\{ \hat{\mu}_i^{\ast,t+1}~\middle|~ 1 \le i \le N^\ast \right\},\\
    \hat{\mu}_i^{\ast,t+1}&=\mu_i^{\ast,t} + M_i^{\ast,t}.
\end{split}
\end{equation}

This decoding step translates the learned latent dynamic representations into explicit motion trajectories. The details of how the 3D positions $\hat{G}_{t+1}$ and 3D rotations of raw particles for the next frame are interpolated from the Key Points $\hat{G}_{t+1}^\ast$ using LBS are provided in the Supplementary Material. Here, LBS serves as a dense motion propagation mechanism that smoothly transfers predicted Key Point motions to raw particles while preserving coherent object deformation and rendering consistency.

\subsection{Training Strategy}

We minimize the Mean Squared Error (MSE) between the final next-frame raw particle positions $\hat{G}_{t+i}$ and the ground truth $G_{t+i}$ over the subsequent $\epsilon$ frames to train DyG$^2$T,
\begin{equation}
\mathcal{L}_{\text{pred}}=\sum_{i=1}^{\epsilon}\left\|\hat{G}_{t+i}-G_{t+i}\right\|^{2}.
\end{equation}

In practice, we set $\epsilon=5$ to provide a short-horizon yet sufficiently informative temporal window. Supervising multiple future frames encourages the DyG$^2$T to generate dynamically stable and drift-resistant dynamics reasoning, rather than overfitting to a single-step displacement.

\section{Experiment}
\subsection{Experiment Settings}\label{sec_Experiment_Settings}

\subsubsection{Dataset}
We evaluate DyG$^2$T on the Spring-Gaus dataset~\cite{zhong2024reconstruction} and our Unity3D-Heterogeneous (Unity3D-H) dataset. Spring-Gaus (synthetic) contains elastic objects with diverse materials and appearances, recorded as 30-frame $512\times512$ videos from 10 views, with MPM-simulated trajectories as 3D ground truth. Its real-world subset provides 20-frame $1920\times1080$ 3 views videos of five toys dropped from rest, together with 50$\sim$70 static multiview images for appearance reconstruction. Unity3D-H is collected using the Unity3D engine~\cite{wang2010new}, featuring a polyhedral object composed of two rigidly connected elastic materials. Released from random upper-hemisphere positions without initial velocity, the object exhibits material-dependent bouncing trajectories. The dynamic sequence is rendered as a 30-frame $2098\times1327$ video from 10 views. All cameras across the above datasets are fixed and uniformly distributed over the upper hemisphere surrounding the dynamic scene. 

Following the Spring-Gaus~\cite{zhong2024reconstruction}, DyG$^2$T learns dynamics from the first 20 frames (10 for real-world Spring-Gaus) of each observation, while the final 10 frames, which are unseen during training, are reserved for dynamic modeling and reasoning evaluation. Additionally, following prior work~\cite{xue20233d, zhong2024reconstruction}, we use GroundingDINO~\cite{liu2024grounding} and SAM~\cite{kirillov2023segment} to extract dynamic-object masks and remove background interference.
 
\subsubsection{Baselines}

Following prior work, we evaluate DyG$^2$T from two perspectives. For dynamic reconstruction, we compare against Spring-Gaus~\cite{zhong2024reconstruction}, which performs per-frame geometry optimization using a spring-mass model initialized from the first frame. For dynamics modeling, we adopt GS-Dynamics~\cite{zhang2025dynamic}, a differentiable network-based pipeline combining Dyn3DGS with GNN-based dynamics prediction, as well as Spring-Gaus as baselines. Furthermore, for the challenging heterogeneous-material dynamics scenario, we additionally introduce 2 strong physics-prior baselines. PAC-NeRF~\cite{li2023pac} characterizes dynamic scenes by integrating neural radiance fields with continuum mechanics modeling, incorporating explicit physical parameters and differentiable simulation constraints. GIC~\cite{cai2024gic}, in contrast, assists object dynamics modeling through physical parameter inversion.

All baselines are re-trained and re-evaluated using the authors’ recommended hyperparameters, and we report their best-performing results. To ensure fairness, both GS-Dynamic and DyG$^2$T perform dynamics modeling using the same Dyn3DGS reconstruction results.

\subsubsection{Metrics}
Following prior work~\cite{zhang2025dynamic}, we adopt CD, the bidirectional L2 distance between predicted and ground-truth point clouds, and EMD, the minimal transport cost between two point sets, as our 3D trajectory evaluation metrics. For 2D appearance assessment, we measure the similarity between rendered and ground-truth images using PSNR, SSIM~\cite{wang2004image}, and LPIPS~\cite{zhang2018unreasonable} across multiple views. PSNR and SSIM quantify pixel-level discrepancies, whereas LPIPS leverages AlexNet-based perceptual features to capture human-aligned differences in overall visual appearance. Because baseline methods do not release metric-evaluation code, we reimplement all metrics in a unified framework to ensure fair comparison. 3D metrics are averaged over all evaluation frames, while 2D metrics are averaged across views per frame and then across all evaluation frames.

\subsubsection{Implementation Details} 

We train DyG$^2$T for 1000 epochs, with each epoch consisting of 100 iterations. The PointNet neighborhood size $k$ is $16$ ($k=8$ for Apple and Toothpaste), and the sharpness parameter $\rho$ is chosen as half of the minimum pairwise distance between Key Points. To increase data diversity and improve robustness to reconstruction noise, we augment each dynamic reconstruction trajectory into 30 instances by adding a uniformly sampled perturbation from $[-0.3, 0.3]$ to the Raw Point Cloud of every frame. The test split uses only the clean, unaugmented reconstructions to ensure unbiased evaluation. All augmented trajectories are divided into training and test sets at a 4:1 ratio. All experiments are conducted with a single NVIDIA RTX 4090 GPU.

\vspace{-0.8em}
\subsection{Dynamic Reconstruction of Dynamic Objects}\label{sec_exp_Dynamic_Reconstruction}

To evaluate the dynamic reconstruction quality in DyG$^2$T, we follow standard practice and report CD and EMD on the recovered 3D trajectories. As shown in Table~\ref{tab_dynamicReconstruction}, our reconstruction module, built upon a strong dynamic 3D Gaussian representation backbone Dyn3DGS~\cite{luiten2024dynamic}, produces accurate and stable 3D trajectories across various objects. 

Although dynamic reconstruction is not the primary focus of this work, its reliability is essential because these trajectories serve as the sole supervision signal for the downstream dynamics modeling. The quantitative results verify that the reconstructed dynamic sequences are sufficiently precise to support the downstream modules. Additional qualitative results are provided in the Supplementary Material.

\vspace{-0.8em}
\subsection{Dynamics Modeling and Reasoning of Dynamic Objects}

\subsubsection{Synthetic Dynamic Scenes}

\begin{figure*}[t]
\centering
\includegraphics[width=0.9\textwidth, trim=0 20 0 20]{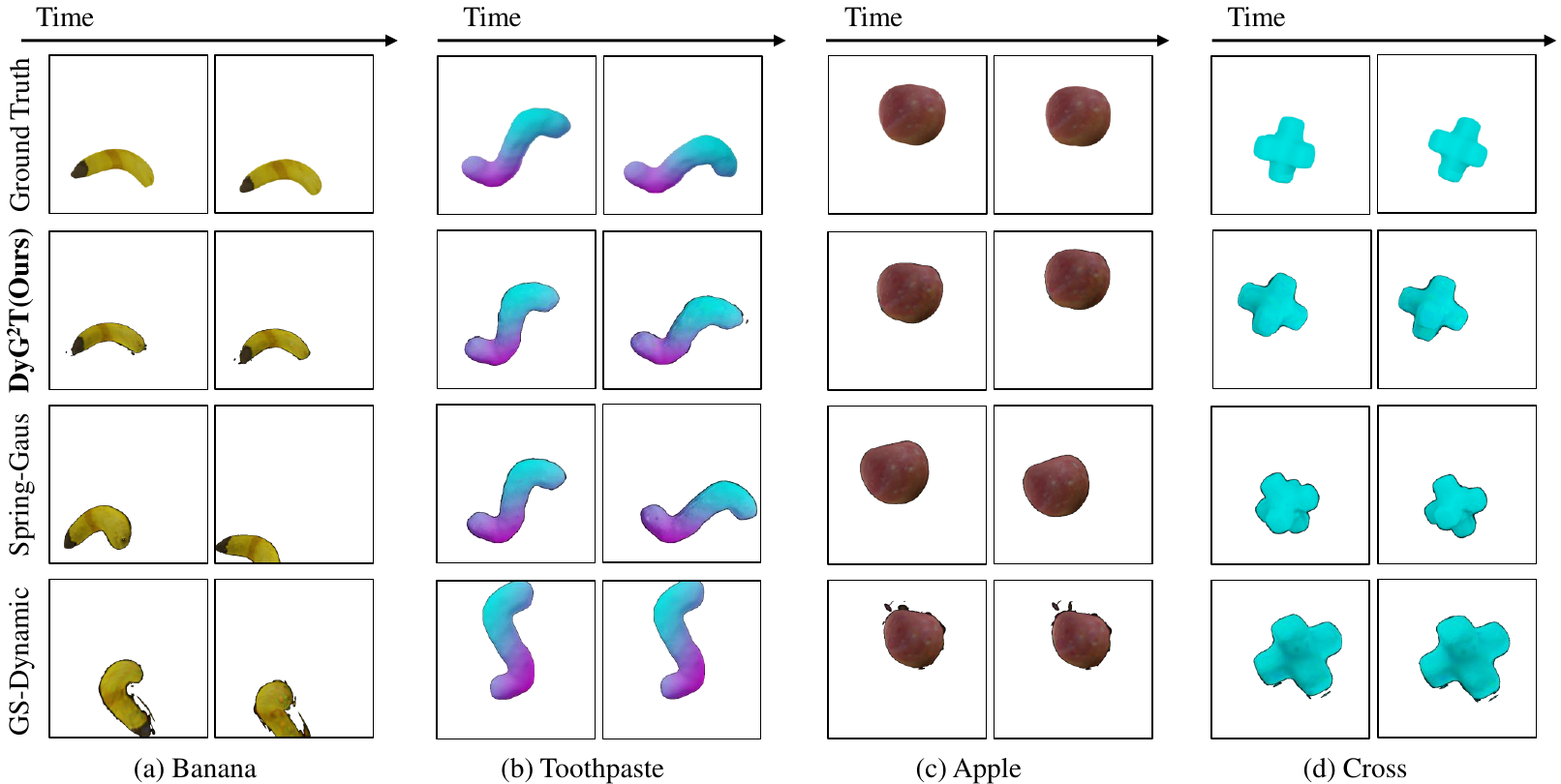}
\caption{Qualitative results of dynamics reasoning. (a), (b), (c), and (d) correspond to Banana, Toothpaste, Apple, and Cross, respectively.}
\label{fig_dynamicsModeling}
\vspace{-1em}
\end{figure*}

To assess dynamics modeling and reasoning, we perform 3D trajectory forecasting on the Spring-Gaus synthetic datasets. As summarized in Table~\ref{tab_motionPrediction}, DyG$^2$T delivers strong performance on this benchmark, especially in 3D trajectory metrics (CD and EMD), demonstrating its ability to capture physically faithful dynamics. While DyG$^2$T ranks second on Cross in PSNR and SSIM, it surpasses all baselines in LPIPS, which more closely reflects human perceptual similarity. Qualitative results in Figure~\ref{fig_dynamicsModeling} further show that predictions from Spring-Gaus~\cite{zhong2024reconstruction} and GS-Dynamics~\cite{zhang2025dynamic} suffer from significant positional drift, whereas DyG$^2$T produces more accurate dynamic trajectories and sharper appearance details, demonstrating strong generalization across diverse object geometries.

\begin{table}
\centering
\tabcolsep=2.6pt
\caption{Quantitative results of dynamic reconstruction for DyG$^2$T and baselines on Spring-Gaus synthetic dataset.}
\label{tab_dynamicReconstruction}
\begin{tabular}{ccccc}
\toprule
\multicolumn{1}{l}{}        & \multicolumn{2}{c}{CD↓}                                               & \multicolumn{2}{c}{EMD↓}                                          \\ \cmidrule{2-5} 
\multicolumn{1}{l}{Objects} & \multicolumn{1}{l}{Spring-Gaus} & \multicolumn{1}{l}{DyG$^2$T(Ours)}     & \multicolumn{1}{l}{Spring-Gaus} & \multicolumn{1}{l}{DyG$^2$T(Ours)} \\ \midrule
\multicolumn{1}{c|}{Torus}  & 0.012                           & \multicolumn{1}{c|}{\textbf{0.008}} & 0.003                           & \textbf{0.001}                  \\
\multicolumn{1}{c|}{Cross}  & 0.016                           & \multicolumn{1}{c|}{\textbf{0.010}} & 0.005                           & \textbf{0.002}                  \\
\multicolumn{1}{c|}{Cream}  & 0.014                           & \multicolumn{1}{c|}{\textbf{0.012}} & 0.007                           & \textbf{0.005}                  \\
\multicolumn{1}{c|}{Apple}  & 0.014                           & \multicolumn{1}{c|}{\textbf{0.011}} & 0.006                           & \textbf{0.003}                  \\
\multicolumn{1}{c|}{Paste}  & 0.011                           & \multicolumn{1}{c|}{\textbf{0.008}} & 0.003                           & \textbf{0.002}                  \\
\multicolumn{1}{c|}{Chess}  & 0.017                           & \multicolumn{1}{c|}{\textbf{0.010}} & 0.007                           & \textbf{0.002}                  \\
\multicolumn{1}{c|}{Banana} & 0.049                           & \multicolumn{1}{c|}{\textbf{0.007}} & 0.027                           & \textbf{0.002}                  \\ \midrule
\multicolumn{1}{c|}{Mean}   & 0.019                           & \multicolumn{1}{c|}{\textbf{0.010}} & 0.008                           & \textbf{0.003}                  \\ \bottomrule
\end{tabular}
\vspace{-1em}
\end{table}

\begin{table*}[t]
\centering
\tabcolsep=11pt
\caption{Quantitative results of dynamics modeling and reasoning on Spring-Gaus synthetic dataset.}
\label{tab_motionPrediction}
\begin{tabular}{c|c|ccccccc|c}
\toprule
          Metrics              & Methods     & Torus           & Cross           & Cream           & Apple           & Paste           & Chess           & Banana          & Mean            \\ \midrule
\multirow{3}{*}{CD↓}    & Spring-Gaus & 0.033           & 0.046           & 0.032           & 0.047           & 0.068           & 0.053           & 0.184           & 0.066           \\
                        & GS-Dynamics & 0.073           & 0.122           & 0.154           & 0.043           & 0.227           & 0.284           & 0.328           & 0.176           \\
                        & DyG$^2$T(Ours) & \textbf{0.029}  & \textbf{0.038}  & \textbf{0.027}  & \textbf{0.028}  & \textbf{0.038}  & \textbf{0.050}  & \textbf{0.055}  & \textbf{0.039}  \\ \midrule
\multirow{3}{*}{EMD↓}   & Spring-Gaus & 0.014           & 0.024           & 0.023           & 0.029           & 0.035           & 0.027           & 0.091           & 0.035           \\
                        & GS-Dynamics & 0.033           & 0.062           & 0.097           & 0.021           & 0.164           & 0.200           & 0.171           & 0.107           \\
                        & DyG$^2$T(Ours) & \textbf{0.013}  & \textbf{0.018}  & \textbf{0.013}  & \textbf{0.015}  & \textbf{0.020}  & \textbf{0.021}  & \textbf{0.029}  & \textbf{0.019}  \\ \midrule
\multirow{3}{*}{PSNR↑}  & Spring-Gaus & 12.220          & \textbf{11.993} & 11.267          & 17.443          & 11.016          & 11.305          & 15.949          & 13.028          \\
                        & GS-Dynamics & 13.450          & 10.621          & 12.647          & 19.632          & 11.506          & 11.758          & 16.622          & 13.748          \\
                        & DyG$^2$T(Ours) & \textbf{14.048} & 11.632          & \textbf{14.765} & \textbf{20.477} & \textbf{14.698} & \textbf{15.653} & \textbf{17.904} & \textbf{15.587} \\ \midrule
\multirow{3}{*}{SSIM↑}  & Spring-Gaus & 0.850           & \textbf{0.876}  & 0.709           & 0.828           & 0.775           & 0.755           & 0.865           & 0.808           \\
                        & GS-Dynamics & 0.876           & 0.842           & 0.763           & 0.887           & 0.802           & 0.749           & 0.880           & 0.828           \\
                        & DyG$^2$T(Ours) & \textbf{0.895}  & 0.871           & \textbf{0.875}  & \textbf{0.907}  & \textbf{0.887}  & \textbf{0.873}  & \textbf{0.919}  & \textbf{0.889}  \\ \midrule
\multirow{3}{*}{LPIPS↓} & Spring-Gaus & 0.349           & 0.303           & 0.370           & 0.230           & 0.332           & 0.335           & 0.250           & 0.310           \\
                        & GS-Dynamics & 0.197           & 0.280           & 0.324           & 0.163           & 0.317           & 0.306           & 0.210           & 0.257           \\
                        & DyG$^2$T(Ours) & \textbf{0.139}  & \textbf{0.220}  & \textbf{0.189}  & \textbf{0.131}  & \textbf{0.178}  & \textbf{0.207}  & \textbf{0.122}  & \textbf{0.171}  \\ \bottomrule
\end{tabular}
\vspace{-0.5em}
\end{table*}

\begin{table*}[t]
\centering
\tabcolsep=11pt
\caption{Quantitative results of dynamics modeling and reasoning on Spring-Gaus real-world dataset and our Unity3D-H dataset.}
\label{tab_motionPrediction2}
\begin{tabular}{cccccccc}
\toprule
\multirow{2}{*}{Metrics}                     & \multirow{2}{*}{Methods}         & \multicolumn{5}{c}{Spring-Gaus real-world}                                                                   & Unity3D-H         \\ \cmidrule{3-8} 
                                             &                                  & Dog             & Potato          & Pig             & Burger          & Bun                                  & Polyhedron      \\ \midrule
\multicolumn{1}{c|}{\multirow{3}{*}{PSNR↑}}  & \multicolumn{1}{c|}{Spring-Gaus} & 21.499          & 20.881          & 21.136          & 21.026          & \multicolumn{1}{c|}{20.456}          & 31.027          \\
\multicolumn{1}{c|}{}                        & \multicolumn{1}{c|}{GS-Dynamics} & 26.141          & \textbf{28.623} & 27.114          & 27.969          & \multicolumn{1}{c|}{26.929}          & 31.352          \\
\multicolumn{1}{c|}{}                        & \multicolumn{1}{c|}{DyG$^2$T(Ours)} & \textbf{27.676} & 27.933          & \textbf{27.750} & \textbf{30.645} & \multicolumn{1}{c|}{\textbf{27.197}} & \textbf{31.768} \\ \midrule
\multicolumn{1}{c|}{\multirow{3}{*}{SSIM↑}}  & \multicolumn{1}{c|}{Spring-Gaus} & 0.987           & 0.985           & 0.986           & 0.985           & \multicolumn{1}{c|}{0.984}           & 0.985           \\
\multicolumn{1}{c|}{}                        & \multicolumn{1}{c|}{GS-Dynamics} & 0.988           & \textbf{0.989}  & 0.989           & 0.988           & \multicolumn{1}{c|}{0.988}           & 0.987           \\
\multicolumn{1}{c|}{}                        & \multicolumn{1}{c|}{DyG$^2$T(Ours)} & \textbf{0.991}  & 0.987           & \textbf{0.989}  & \textbf{0.994}  & \multicolumn{1}{c|}{\textbf{0.988}}  & \textbf{0.990}  \\ \midrule
\multicolumn{1}{c|}{\multirow{3}{*}{LPIPS↓}} & \multicolumn{1}{c|}{Spring-Gaus} & 0.030           & 0.032           & 0.031           & 0.031           & \multicolumn{1}{c|}{0.032}           & 0.020           \\
\multicolumn{1}{c|}{}                        & \multicolumn{1}{c|}{GS-Dynamics} & 0.023           & \textbf{0.020}  & 0.020           & 0.022           & \multicolumn{1}{c|}{0.020}           & 0.018           \\
\multicolumn{1}{c|}{}                        & \multicolumn{1}{c|}{DyG$^2$T(Ours)} & \textbf{0.019}  & 0.021           & \textbf{0.017}  & \textbf{0.012}  & \multicolumn{1}{c|}{\textbf{0.018}}  & \textbf{0.015}  \\ \bottomrule
\end{tabular}
\vspace{-1em}
\end{table*}

\subsubsection{Real-world and Cross-benchmark Generalization}

Furthermore, as shown in Table~\ref{tab_motionPrediction2}, the evaluation on the Spring-Gaus real-world and our Unity3D-H datasets demonstrates that DyG$^2$T exhibits strong real-world generalization and can be readily transferred to other benchmarks. Notably, while GS-Dynamics achieves leading performance on the relatively simple Potato, it falls behind on toys with more complex geometries, such as Burger, where DyG$^2$T's spatiotemporally completed particle features and multi-scale propagation paths provide a clear advantage.

\begin{table}[t]
\centering
\caption{Quantitative results of DyG$^2$T, its variants, and baselines for heterogeneous torus dynamics modeling and reasoning on the Spring-Gaus synthetic dataset.}
\label{tab_heterogeneous}{
\begin{tabular}{c|ccccc}
\toprule
Methods     & CD↓            & EMD↓           & PSNR↑           & SSIM↑          & LPIPS↓         \\ \midrule
Spring-Gaus & 0.030          & 0.012          & 12.197          & 0.850          & 0.350          \\
GS-Dynamics & 0.116          & 0.049          & 13.342          & 0.861          & 0.235          \\
PAC-NeRF    & 0.284          & 0.203          & \textbf{15.188}          & \textbf{0.900}          & 0.187          \\
GIC         & 0.116          & 0.132          & 14.009          & 0.882          & 0.217           \\
DyG$^2$T$_{\text{noisy}}$ & 0.035          & 0.016          & 13.647          & 0.878          & 0.188          \\
DyG$^2$T(Ours) & \textbf{0.015} & \textbf{0.000} & 14.080 & 0.893 & \textbf{0.129} \\ \bottomrule
\end{tabular}
}
\vspace{-1.5em}
\end{table}

Compared with the synthetic and real subsets of Spring-Gaus, the performance gain of DyG$^2$T on Unity3D-H is smaller due to differences in their simulation and observation pipelines. Spring-Gaus synthetic subset, produced via a staged simulate-then-render process, introduces additional noise, while the real subset further suffers from deformation noise and non-ideal physical interactions. These perturbations amplify semantic collapse, thereby accentuating DyG$^2$T’s advantage in modeling multi-scale interaction using temporally discriminative, complete particle features. By contrast, Unity3D-H is generated through a physically consistent one-step simulator, producing cleaner motion and weaker temporal collapse, so the relative improvement appears smaller. Nevertheless, DyG$^2$T achieves stable gains, and Unity3D-H further demonstrates its cross-benchmark generalization.

\begin{figure}[t]
\centering
\includegraphics[width=\columnwidth, trim=0 20 0 0]{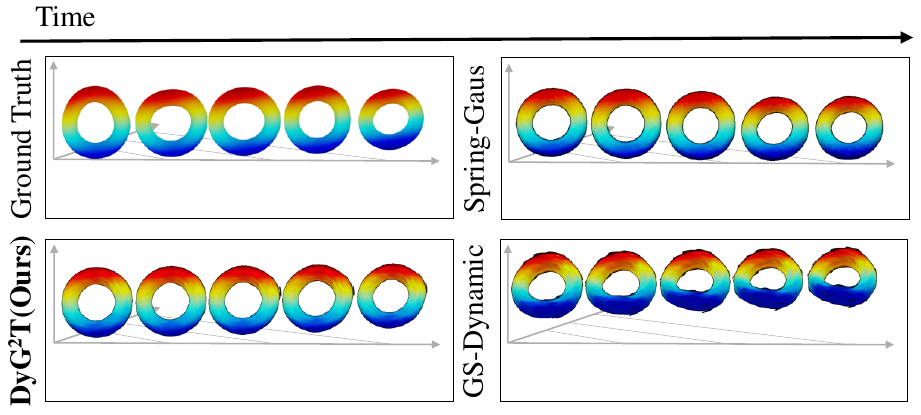}
\caption{Qualitative results of dynamics modeling and reasoning on the Heterogeneous Torus. The spatial positions of dynamic objects are calibrated using a consistent coordinate system.}
\label{fig_heterogeneous}
\vspace{-0.5em}
\end{figure}

\begin{table}[t]
\centering
\tabcolsep=8pt
    \caption{Quantitative results of dynamics modeling and reasoning physical consistency evaluation on the Heterogeneous Torus.}
    \begin{tabular}{cccc}
    \toprule
    Methods     & LSE↓           & SPC↑           & ACE↓           \\ \midrule
    GS-Dynamics & 0.499          & -0.248         & 0.017          \\
    DyG$^2$T(Ours) & \textbf{0.225} & \textbf{0.089} & \textbf{0.016} \\ \bottomrule
    \end{tabular}
    \label{table_phy_eval}
\vspace{-1.5em}
\end{table}

\subsubsection{Robustness to Heterogeneous Objects and Noise Input}

We conduct experiments on a Heterogeneous Torus with complex physical properties. As shown in Table~\ref{tab_heterogeneous} and Figure~\ref{fig_heterogeneous}, DyG$^2$T delivers more accurate trajectory predictions by leveraging spatiotemporally completed particle representations and multi-scale interaction modeling, enabling the model to capture subtle variations in material response. Relative to strong physics-prior approaches such as PAC-NeRF and GIC, DyG$^2$T is still able to preserve 3D geometric distribution quality more accurately, particularly exhibiting more stable dynamics modeling capability in terms of particle trajectories and deformation structures. The results on the Unity3D-H dataset in Table~\ref{tab_motionPrediction2} also validate DyG$^2$T’s scalability to heterogeneous objects across different benchmarks. 

Additionally, we introduce 3 physics-consistency metrics sensitive to heterogeneous objects to further evaluate dynamics reasoning on the Heterogeneous Torus, including Local Stretch Error (LSE), Strain Pattern Correlation (SPC), and Acceleration Consistency Error (ACE). Specifically, LSE measures the consistency of local stretch with Ground Truth, SPC evaluates the spatial correlation of high/low strain regions with Ground Truth, and ACE assesses second-order temporal acceleration consistency with Ground Truth. Detailed formulations are provided in the supplementary material. As shown in Table~\ref{table_phy_eval}, DyG$^2$T achieves superior performance on these metrics, indicating its ability to better preserve local deformation behaviors, accurately capture the spatial distribution of strain intensity, and maintain consistent temporal acceleration.

As shown in Table~\ref{tab_heterogeneous}, we additionally test robustness to imperfect observations by injecting noise into central trajectories and applying non-rigid distortions to the reconstructed point clouds. Thanks to the disentangling and temporal discriminability Key Point features, DyG$^2$T$_{\text{noisy}}$ exhibits only minor degradation in 3D trajectory prediction, while its 2D appearance evaluation still surpasses all baselines.

\begin{figure}[t]
\centering
\includegraphics[width=0.85\columnwidth, trim=0 20 0 0]{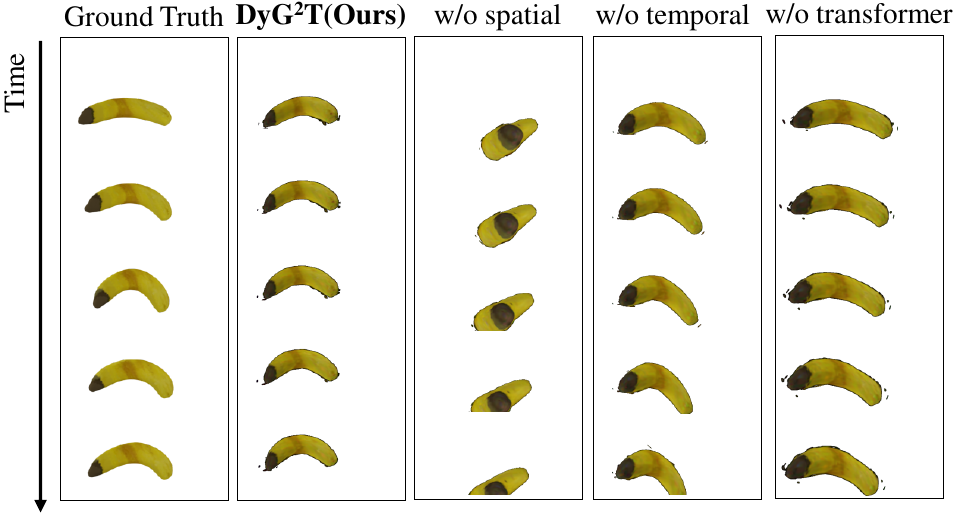}
\caption{Qualitative results of module ablation study. The vertical axis shows the object motion's temporal evolution.}
\label{fig_ablationModules}
\vspace{-1em}
\end{figure}

\begin{table}[t]
\centering
\caption{Quantitative results of module ablation study.}
\label{tab_ablationModules}
\begin{tabular}{c|ccccc}
\toprule
Methods         & CD↓            & EMD↓           & PSNR↑           & SSIM↑          & LPIPS↓         \\ \midrule
w/o spatial     & 0.354          & 0.203          & 16.174          & 0.875          & 0.220          \\
w/o temporal    & 0.227          & 0.120          & 16.493          & 0.882          & 0.198          \\
w/o transformer & 0.197          & 0.110          & 16.264          & 0.875          & 0.201          \\
DyG$^2$T(Ours)     & \textbf{0.055} & \textbf{0.029} & \textbf{17.904} & \textbf{0.919} & \textbf{0.122} \\ \bottomrule
\end{tabular}
\vspace{-1em}
\end{table}

\begin{figure}[t]
\centering
\includegraphics[width=0.9\columnwidth, trim=0 20 0 0]{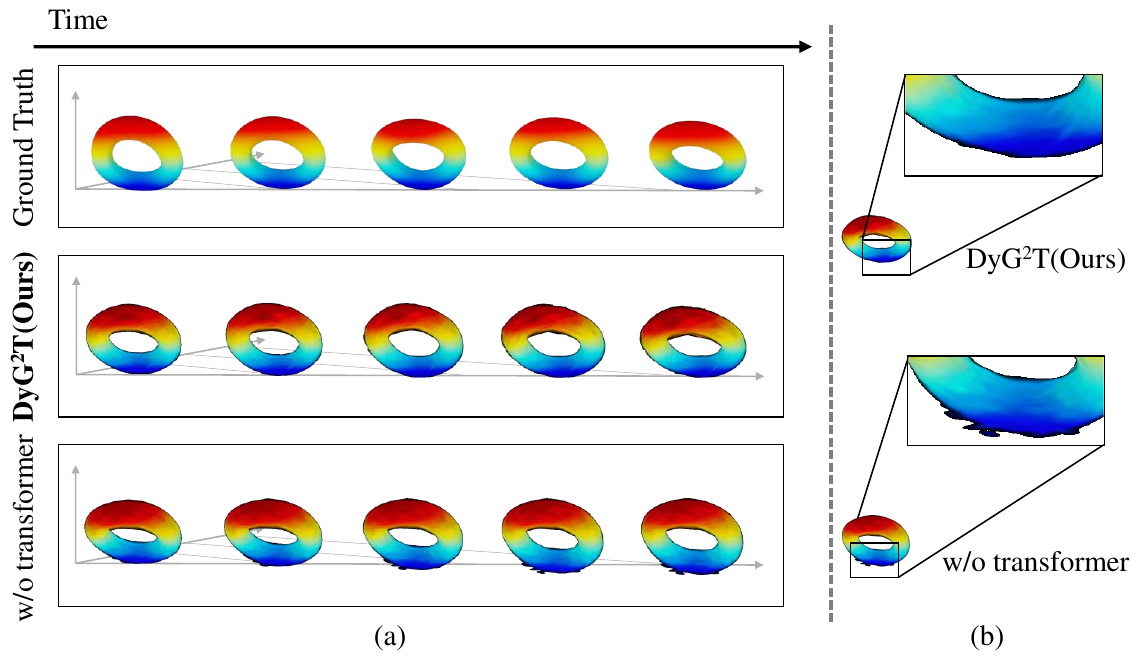}
\caption{Qualitative results of the Particle Graph Transformer ablation study. (a) DyG$^2$T and its w/o transformer variants on the dynamics modeling and reasoning task of the Heterogeneous Torus. (b) Local zoom-in views for evaluating the fine-grained local details.}
\label{fig_ablationModules_GNN}
\vspace{-1em}
\end{figure}

\subsection{Ablation Studies}

To validate the effectiveness of DyG$^2$T’s core components and assess optimal parameter settings, we conduct ablation studies on the Spring-Gaus synthetic dataset.

\subsubsection{The Modules of DyG$^2$T}

The ablation study in Table~\ref{tab_ablationModules} and Figure~\ref{fig_ablationModules} reveals the contribution of each core component in DyG$^2$T. Removing the spatial completion module (w/o spatial), which prevents Key Points from aggregating positional cues from neighboring raw particles and pairwise Key Points, leads to a substantial drop in accuracy. This validates the efficacy of our completed particle representations in restoring fine-grained local and geometric structural cues that sparse Key Point sampling alone cannot preserve.

Eliminating temporal disentangling (TDN) and Temporal Attention (w/o temporal), and instead relying on naive feature concatenation, also results in notable degradation. Without disentangling and temporal discriminability Key Point features, the model fails to maintain coherent cross-frame temporal information, highlighting the critical role of TDN followed by Temporal Attention aggregation.

Finally, replacing the Particle Graph Transformer with a vanilla GNN (w/o transformer) severely limits performance despite having access to the same spatiotemporally completed features. This demonstrates that accurate dynamics modeling hinges on establishing multi-scale interaction paths and capturing long-range dependencies, capabilities that standard local GNNs cannot provide. The qualitative results in Figure~\ref{fig_ablationModules_GNN} further reinforce this conclusion. Due to the heterogeneous material properties of the Torus, the w/o transformer variant accumulates noticeable errors in the later stages of dynamics reasoning (Figure~\ref{fig_ablationModules_GNN}(a)), leading to visible artifacts in the inferred appearance (Figure~\ref{fig_ablationModules_GNN}(b)). 

\subsubsection{Particle-level Spatial Semantic Completion}

\begin{figure*}[t]
\centering
\includegraphics[width=\textwidth, trim=0 20 0 10]{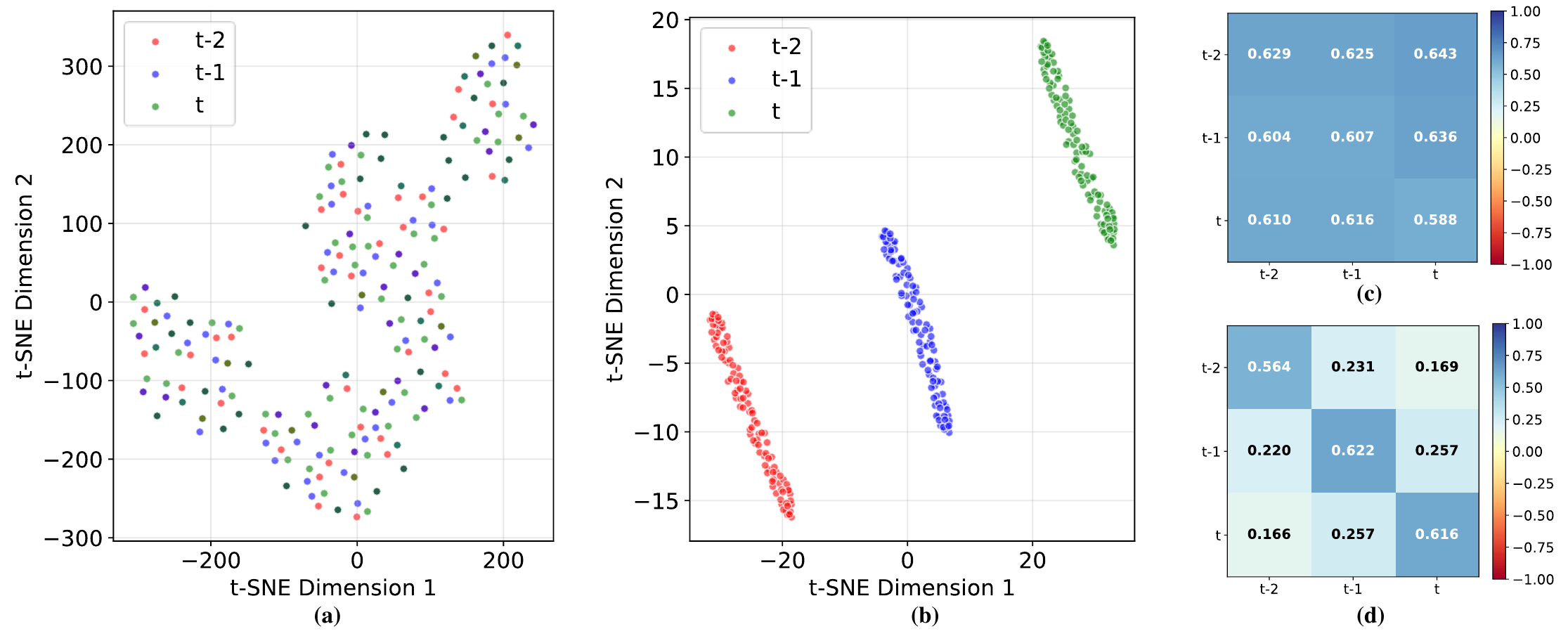}
\caption{Visualization of the Temporal Disentangling Net (TDN) ablation. The Key Point feature embeddings before (a) and after (b) passing through TDN from frames $t-2$, $t-1$, and $t$ are projected into 2D using t-SNE and plotted as scatter plots, with colors indicating frame indices. Cosine similarities among the 3 frames’ Key Point embeddings are also computed before (c) and after (d) TDN and visualized as heatmaps.}
\label{fig_ablationTDN}
\vspace{-1em}
\end{figure*}

To assess the role of spatial particle completion, we vary the neighborhood size $k$ used to aggregate Raw Point features $X_{\text{Po}}^t$. As shown in Table~\ref{tab_ablationSpatial_Temporal} (Row 1\&2), using too small a neighborhood (DyG$^2$T$_{k=8}$) prevents each Key Point from capturing sufficient fine-grained local details from neighboring raw particles, weakening the spatial completion process and causing notable degradation in trajectory prediction. Conversely, an overly large neighborhood (DyG$^2$T$_{k=32}$) introduces redundant or non-informative points, diluting the fine-grained local signals that the spatial module aims to preserve. Empirically, $k=16$ provides the best balance ($k=8$ for Apple and Toothpaste), enabling each Key Point to reconstruct fine-grained local details effectively.

To validate the local Raw Point neighborhood aggregation strategy, we compare max pooling with average pooling (PN Avg) and attention-based pooling (PN Att) while keeping all other settings unchanged. As shown in Table~\ref{tab_ablationSpatial_Temporal} (Row 3\&4), max pooling (Row 11) achieves the best performance across all metrics. This indicates that max pooling provides a more discriminative and noise-robust summary of local Raw Point responses for DyG$^2$T.

We further ablate the components within Particle-level Spatial Semantic Completion. As shown in Table~\ref{tab_ablationSpatial_Temporal} (Row 5\&6), removing the PosDiff Encoder (w/o PosDiff) leads to the largest degradation, confirming that explicitly encoding pairwise Key Point offsets is essential for reconstructing accurate geometric structure. Without these relative positional cues, the model loses its ability to anchor local detail features to the correct spatial layout, severely impairing interaction modeling and trajectory prediction. Similarly, the w/o PosAware variant demonstrates that omitting the Multi-head Position-Aware Attention weakens the alignment between coordinate embeddings and neighborhood features. 

\begin{table}[t]
\centering
\caption{Ablation study on neighborhood ranges $k$ (Row 1\&2), Raw Points neighborhood aggregation strategy (Row 3\&4), the Particle-level Spatial Semantic Completion (Row 5\&6), and the Object-level Dynamic Temporal Aggregation (Row 7$\sim$10).}
\label{tab_ablationSpatial_Temporal}
\begin{tabular}{Nl|ccccc}
\toprule
& Methods      & CD↓            & EMD↓           & PSNR↑           & SSIM↑          & LPIPS↓         \\ \midrule
 & DyG$^2$T$_{k=8}$      & 0.244          & 0.125          & 16.368          & 0.878          & 0.207          \\
 & DyG$^2$T$_{k=32}$      & 0.169          & 0.092          & 16.365          & 0.879          & 0.190          \\ \midrule
 & PN Avg      & 0.107          & 0.240          & 16.071          & 0.871          & 0.242          \\
 & PN Att       & 0.116          & 0.246          & 16.012          & 0.869          & 0.246          \\
\midrule
 & w/o PosDiff  & 0.372          & 0.178          & 16.742          & 0.886          & 0.195          \\
 & w/o PosAware & 0.107          & 0.056          & 17.613          & 0.912          & 0.139          \\ \midrule
 & w/o TDN & 0.107          & 0.240          & 16.071          & 0.871          & 0.242          \\
 &  LSTM        & 0.117          & 0.055          & 17.002          & 0.893          & 0.166          \\
 & avg pool    & 0.096          & 0.052          & 17.344          & 0.906          & 0.145          \\
 & max pool    & 0.365          & 0.195          & 16.290          & 0.874          & 0.221          \\ \midrule
 & DyG$^2$T(Ours)    & \textbf{0.055} & \textbf{0.029} & \textbf{17.904} & \textbf{0.919} & \textbf{0.122} \\ \bottomrule
\end{tabular}
\vspace{-1em}
\end{table}

\begin{table}[t]
    \centering
    \renewcommand{\arraystretch}{1}
    \setlength{\tabcolsep}{1.5mm}
    \caption{Quantitative results of dynamics modeling and reasoning under different observation window sizes.}
    \begin{tabular}{cccccc}
    \toprule
Obs.   Windows & CD↓            & EMD↓           & PSNR↑           & SSIM↑          & LPIPS↓         \\ \midrule
5              & \textbf{0.018} & 0.090          & 16.709          & 0.887          & 0.186          \\
3       & 0.055          & \textbf{0.029} & \textbf{17.904} & \textbf{0.919} & \textbf{0.122} \\ \bottomrule
\end{tabular}
    \label{table_3vs5Frame}
    \vspace{-1em}
\end{table}

\begin{table}[t]
    \centering
    \renewcommand{\arraystretch}{1}
    \setlength{\tabcolsep}{3mm}
    \caption{Quantitative results of dynamics modeling and reasoning on collision and non-collision frames for DyG$^2$T.}
\begin{tabular}{cccc}
\toprule
Frames          & PSNR↑           & SSIM↑          & LPIPS↓         \\ \midrule
Collision Frame & \textbf{18.410} & \textbf{0.927} & \textbf{0.109} \\
Rest Frame      & 17.687          & 0.916          & 0.128    \\ \bottomrule     
\end{tabular}
    \label{table_colFrames}
    \vspace{-1em}
\end{table}

\begin{table}[t]
\centering
\caption{Ablation study of the number of Key Points $N^\ast$ in the Particle Graph.}
\label{tab_ablationGraph_N}
\begin{tabular}{l|ccccc}
\toprule
Variants      & CD↓            & EMD↓           & PSNR↑           & SSIM↑          & LPIPS↓         \\ \midrule
$N^\ast = 50$  & 0.470          & 0.247          & 16.003          & 0.869          & 0.248          \\
$N^\ast = 100$  & \textbf{0.055} & \textbf{0.029} & \textbf{17.904}          & \textbf{0.919} & \textbf{0.122} \\
$N^\ast = 150$  & 0.098          & 0.050          & 17.509          & 0.909          & 0.151          \\ \bottomrule
\end{tabular}
\vspace{-1em}
\end{table}

\begin{table}[t]
\centering
\caption{Ablation study of the presence of edges between Key Points and their top-$k_G$ nearest neighbors.}
\label{tab_ablationGraph_topk}
\begin{tabular}{l|ccccc}
\toprule
Variants      & CD↓            & EMD↓           & PSNR↑           & SSIM↑          & LPIPS↓         \\ \midrule
$k_G=2$ & 0.564          & 0.255          & \textbf{18.436} & 0.910          & 0.183          \\
$k_G=5$  & \textbf{0.055} & \textbf{0.029} & 17.904          & \textbf{0.919} & \textbf{0.122} \\
$k_G=10$  & 0.186          & 0.107          & 16.398          & 0.881          & 0.193          \\ \bottomrule
\end{tabular}
\vspace{-1em}
\end{table}

\subsubsection{Object-level Dynamic Temporal Aggregation}

We ablate the Temporal Disentangling Net (TDN) to examine its role in mitigating cross-frame feature collapse and producing temporally discriminative Key Point representations. As shown in Table~\ref{tab_ablationSpatial_Temporal} (Row 7$\sim$10), removing TDN (w/o TDN) leads to a drop in prediction accuracy. Figure~\ref{fig_ablationTDN} provides a deeper analysis. The t-SNE visualizations in (a) and (b) show that before TDN, Key Point embeddings from frames $t-2$, $t-1$, and $t$ collapse into entangled clusters, evidencing the temporal semantic drift. In contrast, TDN effectively canonicalizes these representations into 3 clearly separated, temporally ordered manifolds, demonstrating successful disentangling. Similarly, the cosine similarity heatmaps in (c) and (d) reveal that TDN substantially reduces spurious cross-frame correlations while strengthening intra-frame consistency, precisely matching the goal of temporally discriminative features. We also compare our Temporal Attention aggregation with LSTM and pooling-based baselines. While these alternatives either blur temporal distinctions (pooling) or accumulate redundant correlations (LSTM), Temporal Attention leverages disentangled TDN features to focus on motion-relevant temporal cues, achieving the best performance.

To further examine the stability of TDN, we extend the observation window from 3 to 5 frames while still using the temporal center as the reference frame to preserve symmetric local disentangling; the ablation study on selecting the center frame as the reference is provided in the Supplementary Material. As shown in Table~\ref{table_3vs5Frame}, the results indicate that the default 3-frame setting remains more effective, indicating that TDN is better suited to local temporal canonicalization, whereas a longer window introduces larger cross-frame discrepancy and makes disentangling less stable. We further partition the evaluation frames into collision and non-collision subsets and report 2D rendering metrics separately. The results in Table~\ref{table_colFrames} show that DyG$^2$T maintains strong predictive quality even on collision frames, suggesting that the proposed temporal disentangling remains effective under abrupt contact-induced motion changes. The slightly better performance on collision frames is likely due to their stronger and more coherent short-term motion cues, whereas non-collision frames often contain weaker residual deformations that are more susceptible to rollout accumulation errors.

\subsubsection{Dynamics Modeling Based on Particle Graph Transformer}

Table~\ref{tab_ablationGraph_N} presents the impact of the number of Key Points $N^\ast$ on dynamics modeling. A small Graph ($N^\ast = 50$) including sparse Key Points, which fail to provide sufficient support for dynamics modeling. Moreover, a large Graph ($N^\ast = 150$) introduces excessive redundancy, which can hinder effective modeling of dynamics. Table~\ref{tab_ablationGraph_topk} investigates the effect of different edge sparsity levels in the Particle Graph. A sparse Particle Graph ($k_G = 2$) lacks sufficient alternative paths for interaction modeling, while a dense graph ($k_G = 10$) increases the difficulty of identifying optimal interaction. Notably, the $k_G = 2$ variant exhibits poor trajectory prediction, causing the rendering images to be filled with a large number of invalid pixels and resulting in abnormally high PSNR; the LPIPS and SSIM still demonstrate the superiority of our method. Accordingly, we set $N^\ast=100,~k_G=5$ for the Spring-Gaus synthetic subset as the best practice, while using $N^\ast=120,~k_G=7$ for the Spring-Gaus real-world and Unity3D-H datasets due to their differing data distributions.

\begin{table}[t]
\centering
\caption{Ablation study of the size of the training temporal window $\epsilon$.}
\label{tab_ablationEpsilon}
\begin{tabular}{c|ccccc}
\toprule
$\epsilon$      & CD↓            & EMD↓           & PSNR↑           & SSIM↑          & LPIPS↓         \\ \midrule
$\epsilon=2$ & 0.106          & 0.056          & 17.047 & 0.897          & 0.161          \\
$\epsilon=5$  & \textbf{0.055} & \textbf{0.029} & 17.904          & 0.919 & \textbf{0.122} \\
$\epsilon=8$  & 4.844          & 2.406          & \textbf{19.062}          & \textbf{0.933}          & 0.143          \\ \bottomrule
\end{tabular}
\vspace{-1em}
\end{table}

\subsubsection{Training Strategy}

We analyze the impact of the prediction window $\epsilon$ in Table~\ref{tab_ablationEpsilon}. A short horizon ($\epsilon=2$) provides limited temporal context and yields unstable long-range dynamics. Extending to $\epsilon=5$ achieves the best overall performance, as supervising multiple future frames encourages DyG$^2$T to capture coherent interaction modeling while keeping gradients stable. Interestingly, $\epsilon=8$ leads to much worse CD/EMD but the highest PSNR/SSIM. Upon inspection, long-horizon predictions cause the object to drift outside the camera frustum, resulting in near-empty frames. Pixel-based metrics are thus inflated by large background regions, whereas 3D-based metrics correctly reveal the physical errors. The optimal setting $\epsilon=5$ is shared across all datasets.

\subsubsection{Efficiency Analysis}

To analyze computational cost, we report inference time (Time), throughput (FPS), parameter count (Params), and FLOPs of DyG$^2$T and its variants on a NVIDIA RTX 4090 (24GB) GPU. As shown in Table~\ref{table_infer_time}, although DyG$^2$T has higher Params and FLOPs than GS-Dynamics, it achieves significantly better runtime efficiency. This is because DyG$^2$T performs particle representation enrichment only on sparse Key Points, using highly parallelizable and GPU-friendly operations such as batched MLPs and attention~\cite{wang2021empirical,segura2023irregular}. Ablation results further identify the source of overhead. Removing the Spatial Semantic Completion module (w/o spatial) leads to substantial reductions in both Params and FLOPs, indicating that most computational cost arises from raw particle aggregation and Key Point directed relation modeling. In contrast, removing the Dynamic Temporal Aggregation (w/o temporal) and Particle Graph Transformer (w/o transformer) results in minor reductions, suggesting a relatively small contribution to overall complexity.

\begin{table}[t]
    \centering
    \tabcolsep=4pt
    \caption{Inference-phase computational resource statistics.}
    \begin{tabular}{ccccc}
    \toprule
    Methods         & Time(s)↓ & FPS↑   & Params(M)↓ & FLOPs(M)↓ \\ \midrule
    GS-Dynamics     & 6.852    & 4.378  & 2.900      & 1707.674  \\
    w/o spatial     & 2.823    & 10.627 & 2.902      & 2884.713  \\
    w/o temporal    & 2.887    & 10.391 & 4.878      & 4410.182  \\
    w/o transformer & 2.862    & 10.482 & 4.647      & 4107.910  \\
    DyG$^2$T~(Ours)     & 3.074    & 9.761  & 4.650      & 4393.837  \\ \bottomrule
    \end{tabular}
    \label{table_infer_time}
    \vspace{-1em}
\end{table}

\section{Limitation}

Despite DyG$^2$T performing effectively in modeling foreground object dynamics and predicting dense Gaussian trajectories from sparse-view observations under fixed point-light illumination, several issues remain.
\begin{itemize}
    \item DyG$^2$T adopts the LBS-based densification that propagates Key Point motions to raw particles through smooth weighted interpolation, which may attenuate deformations at spatial scales smaller than the Key Point spacing, such as localized wrinkles or fine surface details.
    \item DyG$^2$T reuses SH-based appearance descriptors across future frames without explicitly modeling time-varying or view-dependent appearance changes, which may reduce visual fidelity under strong specular reflections or dynamic illumination.
\end{itemize}

\section{Conclusion}

This paper presents DyG$^2$T, a dynamics modeling framework that combines spatiotemporally completed particle representations with multi-scale interaction modeling. DyG$^2$T enriches Key Point features with local details from neighboring raw particles and relative offsets among Key Points, while a Temporal Disentangling Net and Temporal Attention capture frame-wise temporal evolution cues. A Particle Graph Transformer further captures long-range interactions via multi-scale interaction modeling, enabling accurate dynamics prediction. Extensive experiments on synthetic and real-world datasets demonstrate precise trajectory prediction and strong generalization, with ablations validating each component. Future work includes augmenting the current LBS-based densification with residual deformation modeling or particle-wise refinement to better capture localized high-frequency surface variations, and extending the framework with time-varying or view-conditioned appearance residual SH modeling to enhance fidelity under dynamic lighting or specular highlights.


\bibliographystyle{IEEEtran}
\bibliography{IEEEabrv}

\vspace{-20pt}

\begin{IEEEbiography}[{\includegraphics[width=1in,height=1.25in,clip,keepaspectratio]{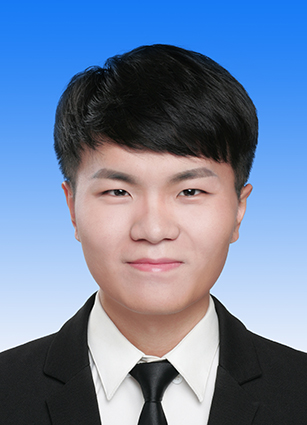}}]{Yansong Wang} received the B.S. degree from Harbin Institute of Technology, Weihai, China, in 2022. He is currently pursuing the Ph.D. degree in computer science and technology at the Harbin Institute of Technology, China. His research interests include computer vision, physical scene understanding, and video understanding.
\end{IEEEbiography}


\begin{IEEEbiography}[{\includegraphics[width=1in,height=1.25in,clip,keepaspectratio]{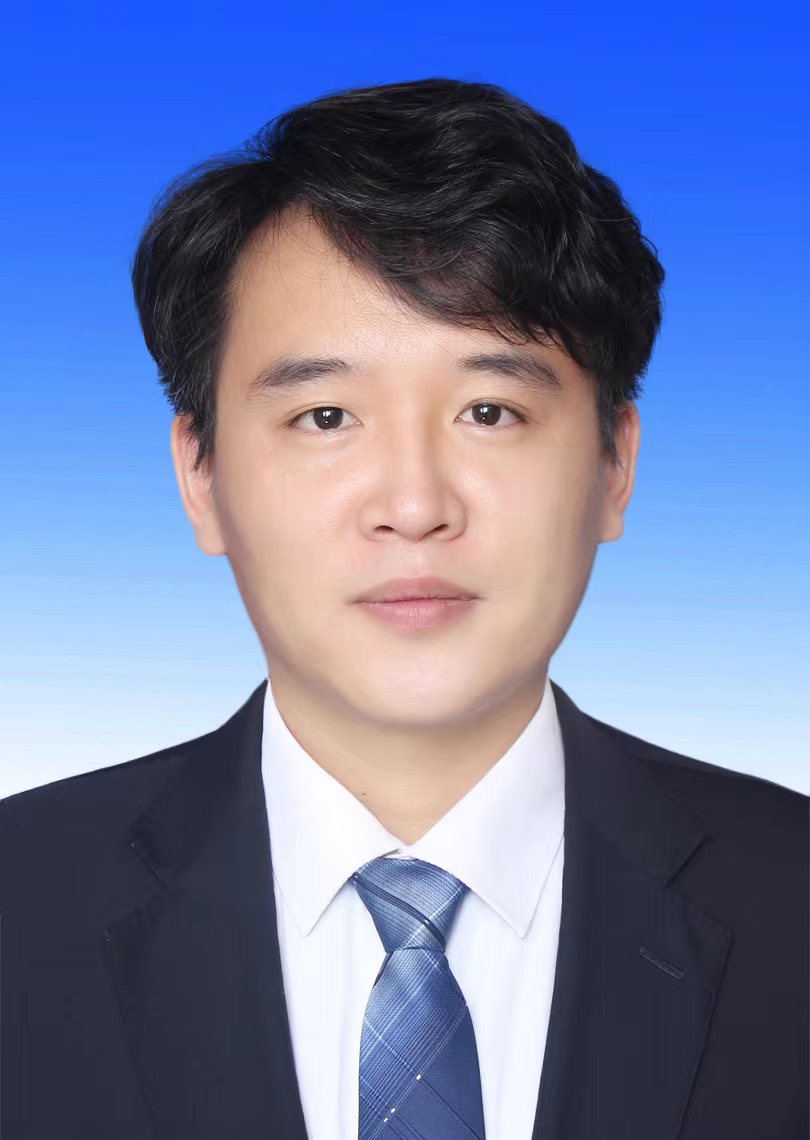}}]{Zhaobo Qi} received the B.S. degree from Harbin Institute of Technology, Weihai, China, in 2016, and the Ph.D. degree from the School of Computer Science and Technology, University of Chinese Academy of Sciences, Beijing, China, in 2022. He is currently an associate professor at Harbin Institute of Technology, Weihai, China. His current research interests include video understanding, knowledge engineering, and computer vision.
\end{IEEEbiography}

\vspace{-5pt}

\begin{IEEEbiography}[{\includegraphics[width=1in,height=1.25in,clip,keepaspectratio]{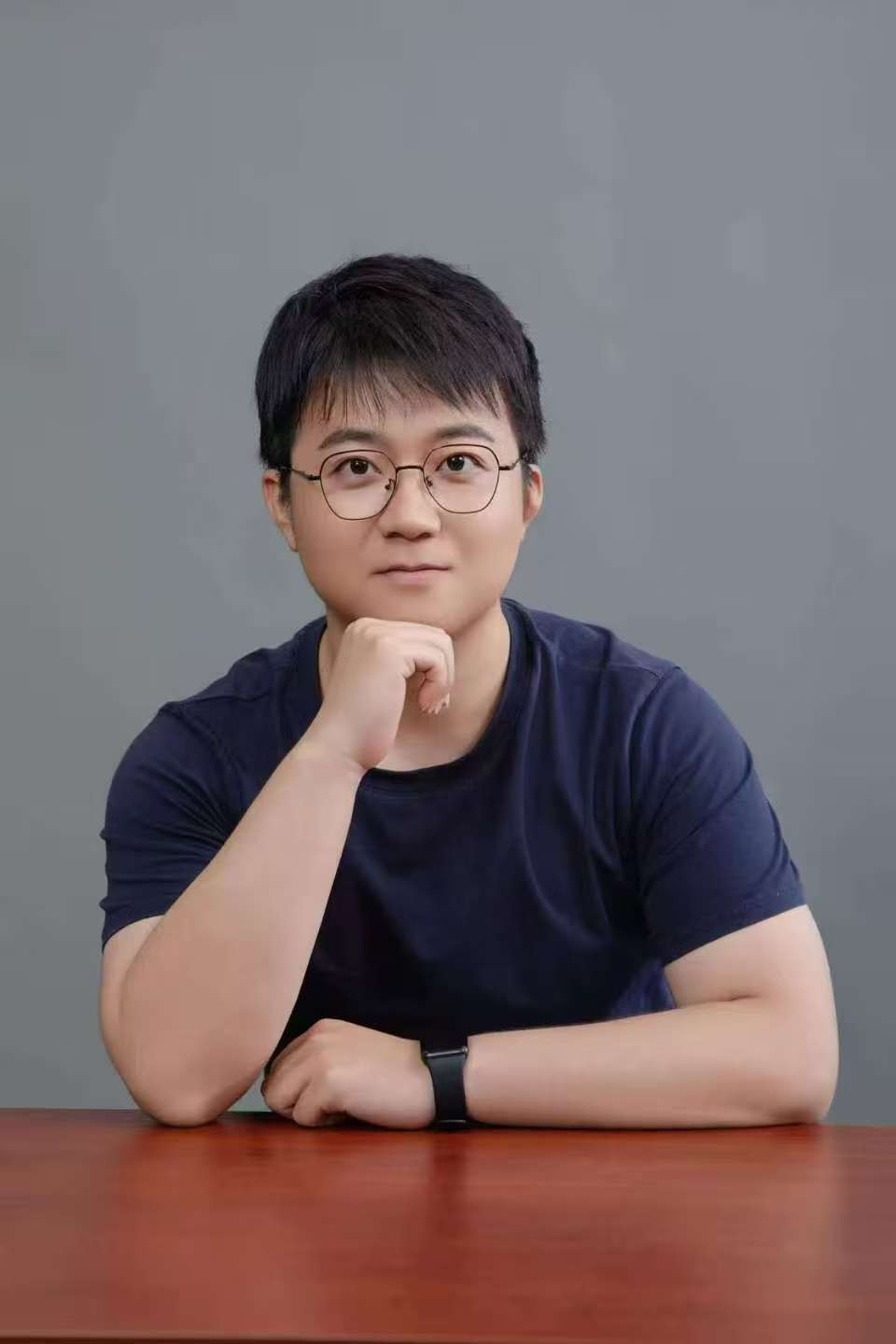}}]{Xinyan Liu} received the Ph.D. degree from the University of Chinese Academy of Sciences, Beijing, China, in 2025. He is currently an assistant professor with the School of Computer Science and Technology, Harbin Institute of Technology, Weihai, China. His research interests include crowd counting and localization, video language tracking.\end{IEEEbiography}


\begin{IEEEbiography}[{\includegraphics[width=1in,height=1.25in,clip,keepaspectratio]{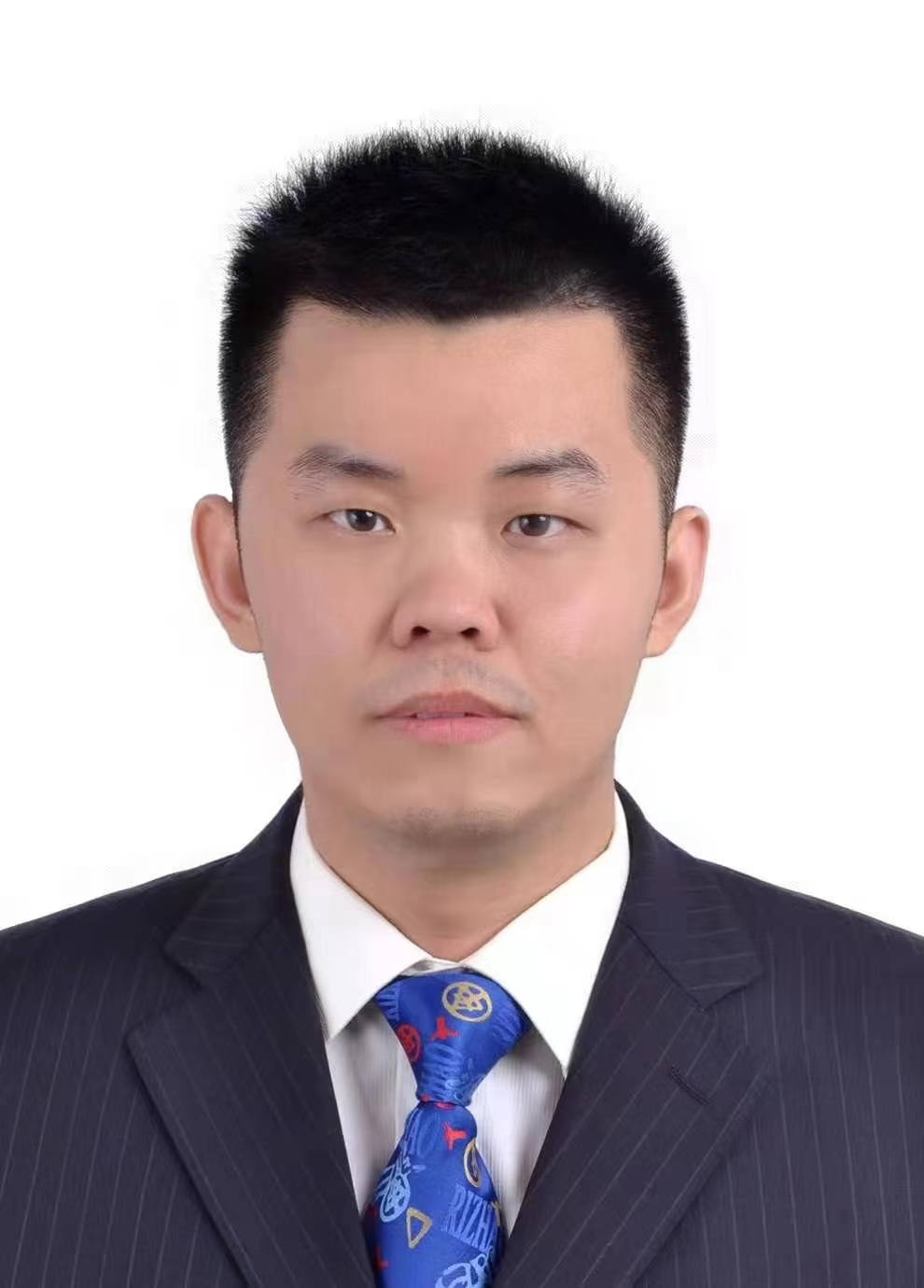}}]{Beichen Zhang} received the B.S. degree and the Ph.D. degree from the University of Chinese Academy of Sciences, China, in 2018 and 2024, respectively. He is currently an associate professor at Harbin Institute of Technology, Weihai. His research interests include active learning, federated learning, and multimedia analysis.\end{IEEEbiography}


\begin{IEEEbiography}[{\includegraphics[width=1in,height=1.25in,clip,keepaspectratio]{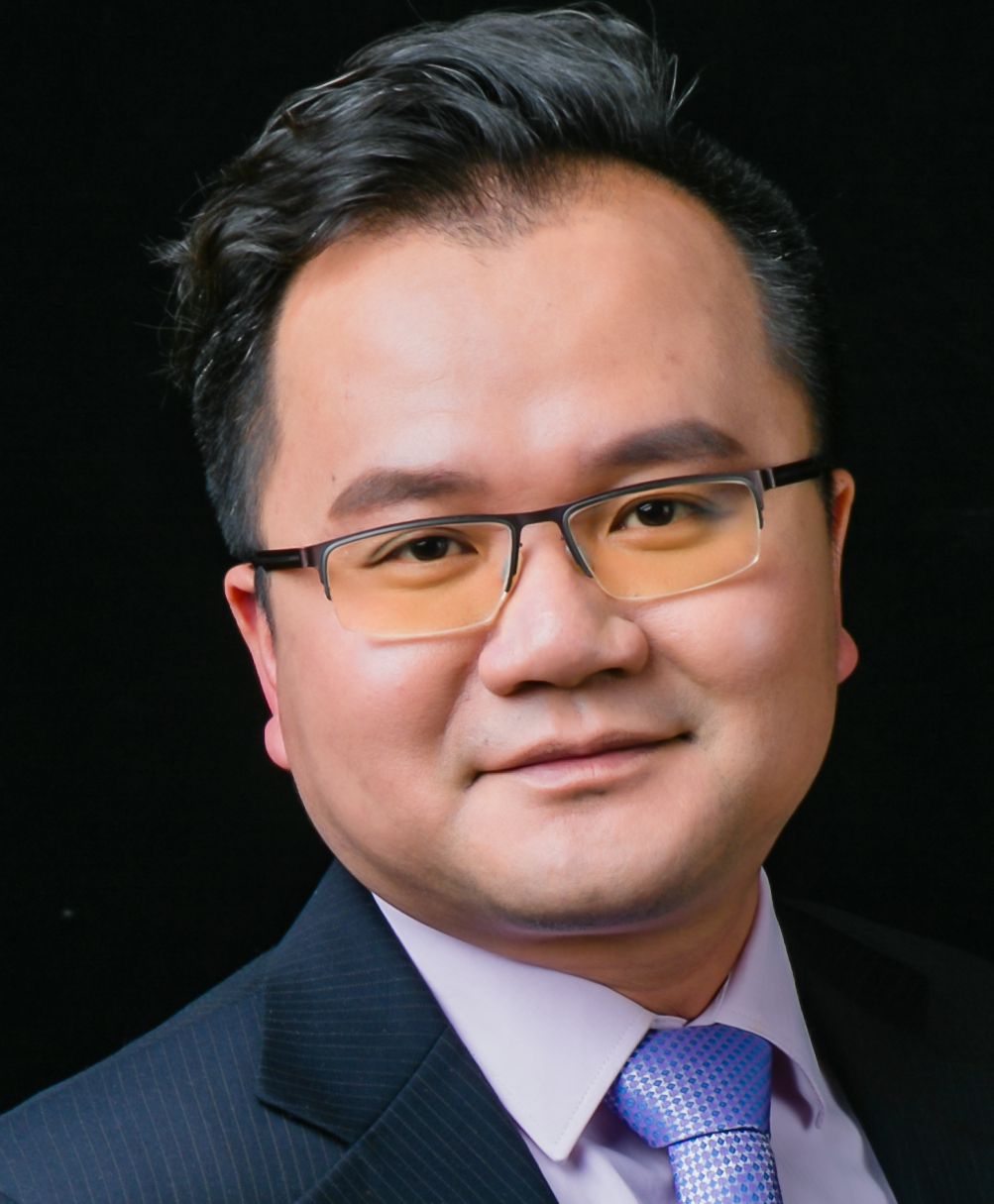}}]{Shuhui Wang} received the B.S. degree in electronics engineering from Tsinghua University, Beijing, China, in 2006, and the Ph.D. degree from the Institute of Computing Technology, Chinese Academy of Sciences, Beijing, China, in 2012. He is currently a Full Professor with the Institute of Computing Technology, Chinese Academy of Sciences. He is also with the Key Laboratory of Intelligent Information Processing, Chinese Academy of Sciences. His research interests include image/video understanding/retrieval, cross-media analysis, and visual-textual knowledge extraction.
\end{IEEEbiography}


\begin{IEEEbiography}[{\includegraphics[width=1in,height=1.25in,clip,keepaspectratio]{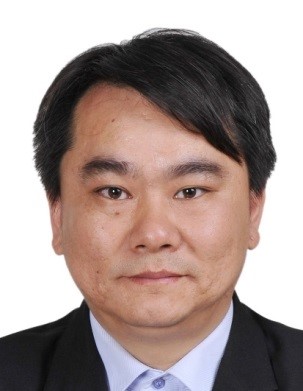}}]{Weigang Zhang} received the B.S. degree in Computer Science and Technology in 2003, the M.S. and Ph.D. degree in Computer Applied Technology in 2005 and 2016, respectively, from Harbin Institute of Technology, Harbin, China. He is currently a Full Professor with the School of Computer Science and Technology, Harbin Institute of Technology, Weihai, China. His research interests include multimedia analysis, computer vision, and pattern recognition. He has published more than 80 academic papers and is the recipient of the Best Student Paper Award at IEEE MIPR 2018.
\end{IEEEbiography}

\begin{IEEEbiography}[{\includegraphics[width=1in,height=1.25in,clip,keepaspectratio]{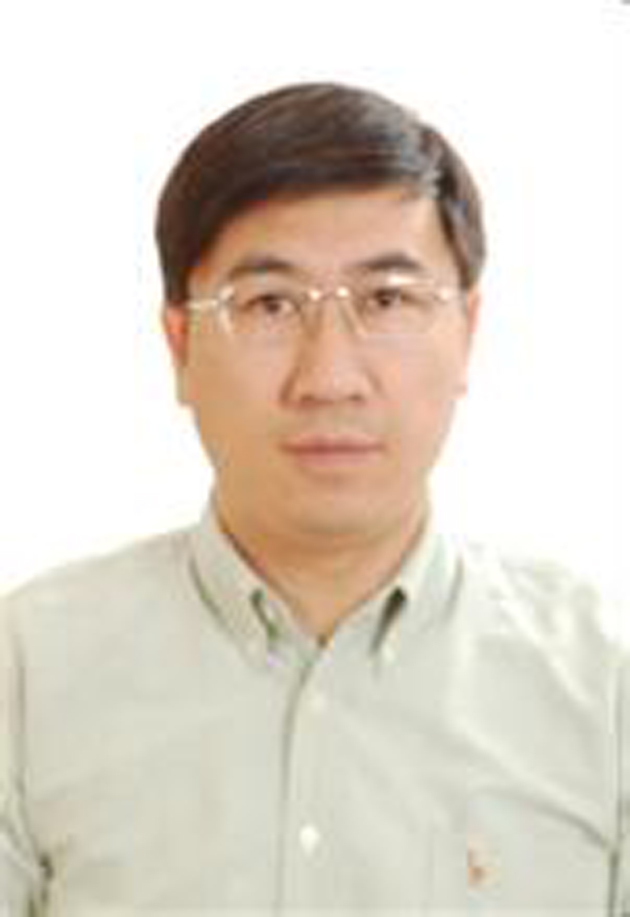}}]{Qingming Huang} received the B.S. degree in computer science and Ph.D. degree in computer engineering from the Harbin Institute of Technology, Harbin, China, in 1988 and 1994, respectively. He is currently a Chair Professor with the School of Computer Science and Technology, University of Chinese Academy of Sciences. He has authored over 400 academic papers in international journals, such as IEEE Transactions on Pattern Analysis and Machine Intelligence, IEEE Transactions on Image Processing, IEEE Transactions on Multimedia, IEEE Transactions on Circuits and Systems for Video Technology, and top level international conferences, including the ACM Multimedia, ICCV, CVPR, ECCV, VLDB, and IJCAI. His research interests include multimedia computing, image/video processing, pattern recognition, and computer vision.
\end{IEEEbiography}

\end{document}